%% file: HY-Video.tex
\documentclass{article}

\PassOptionsToPackage{numbers, compress}{natbib}

\usepackage[final]{neurips_2024}

\usepackage[utf8]{inputenc} 
\usepackage[T1]{fontenc}    
\usepackage{url}            
\usepackage{booktabs}       
\usepackage{amsfonts}       
\usepackage{nicefrac}       
\usepackage{microtype}      
\usepackage{xcolor}         
\usepackage{url}
\usepackage{graphicx}
\usepackage{animate}
\usepackage{listings}
\usepackage{algorithm}
\usepackage{algpseudocode}
\usepackage{enumitem}
\usepackage{silence}
\usepackage{makecell}
\usepackage{graphicx}  
\usepackage{wrapfig}   
\usepackage{caption}
\usepackage{subcaption}
\usepackage{float}
\usepackage{multirow}
\usepackage[accsupp]{axessibility}  
\newcommand{\nameofmethod}{HunyuanVideo}

\newcommand{\myPara}[1]{\noindent\textbf{#1}}

\usepackage[pagebackref,breaklinks,colorlinks,citecolor=gray]{hyperref}

\title{\nameofmethod{}: A Systematic Framework For Large Video Generative Models}

\newif \ifhq
\hqfalse

\begin{document}

\maketitle

\vspace{-20mm}
\begin{quote}
\vspace{-1mm}
\author{Hunyuan Foundation Model Team}
    ``\textit{Bridging the gap between closed-source and open-source video foundation models to accelerate community exploration.}'' \hfill --- \textbf{Hunyuan Foundation Model Team }
\vspace{-1mm}
\end{quote}

\begin{abstract}
Recent advancements in video generation have profoundly transformed daily life for individuals and industries alike. However, the leading video generation models remain closed-source, creating a substantial performance disparity in video generation capabilities between the industry and the public community. 
In this report, we present \nameofmethod{}, a novel open-source video foundation model that exhibits performance in video generation that is comparable to, if not superior to, leading closed-source models. \nameofmethod{} features a comprehensive framework that integrates several key contributions, including data curation, advanced architecture design, progressive model scaling and training, and an efficient infrastructure designed to facilitate large-scale model training and inference.
With those, we successfully trained a video generative model with over 13 billion parameters, making it the largest among all open-source models. 
We conducted extensive experiments and implemented a series of targeted designs to ensure high visual quality, motion dynamics, text-video alignment, and advanced filming techniques. According to professional human evaluation results, \nameofmethod{} outperforms previous state-of-the-art models, including Runway Gen-3, Luma 1.6, and 3 top performing Chinese video generative models. By releasing the code of the foundation model and its applications, we aim to bridge the gap between closed-source and open-source communities. This initiative will empower everyone in the community to experiment with their ideas, fostering a more dynamic and vibrant video generation ecosystem. The code is publicly available at \url{https://github.com/Tencent/HunyuanVideo}.
\end{abstract}

\begin{figure}[!h]
\vspace{-0.5cm}
    \centering
    \begin{minipage}[b]{0.3\textwidth}
        \centering
        \ifhq
        \animategraphics[width=\textwidth]{24}{./hqvideos/teaser/walking_woman/}{0}{128}
        \else
        \animategraphics[width=\textwidth]{24}{./videos/teaser/walking_woman/}{0}{24}
        \fi
    \end{minipage}
    \scalebox{1.03}{
    \begin{minipage}[b]{0.5\textwidth}
        \centering
        \begin{minipage}[t]{0.48\textwidth}
            \centering
            \ifhq
            \animategraphics[width=\textwidth]{24}{./hqvideos/teaser/girl/}{0}{128}
            \else
            \animategraphics[width=\textwidth]{30}{./videos/teaser/girl/}{52}{82}
            \fi
        \end{minipage}%
        \hspace{0.01\textwidth} 
        \begin{minipage}[t]{0.48\textwidth}
            \centering
            \ifhq
            \animategraphics[width=\textwidth]{24}{./hqvideos/teaser/giraffe/}{0}{128}
            \else
            \animategraphics[width=\textwidth]{24}{./videos/teaser/giraffe/}{0}{24}
            \fi
        \end{minipage}
        \begin{minipage}[b]{0.98\textwidth}
            \ifhq
            \animategraphics[width=\textwidth]{24}{./hqvideos/teaser/glass/}{0}{128}
            \else
            \animategraphics[width=\textwidth]{24}{./videos/teaser/glass/}{50}{74}
            \fi
        \end{minipage}
    \end{minipage}}
    \caption{Non-curated multi-ratio generation samples with \nameofmethod{}, showing realistic, concept generalization and automatic scene-cut features.}
\end{figure}

\section{Introduction}
\label{sec:intro}

With extensive pre-training and advanced architectures, diffusion models~\cite{li2024hunyuandit,peebles2023scalable,esser2024scaling,rombach2022high,blattmann2023stable,girdhar2023emu,polyak2024movie,FLUX} have demonstrated superior performance in generating high-quality images and videos compared to previous generative adversarial network (GAN) methods \citep{brock2018large}.
However, unlike the image generation field, which has seen a proliferation of novel algorithms and applications across various open platforms, diffusion-based video generative models remain relatively inactive. We contend that one of the primary reasons for this stagnation is the lack of robust open-source foundation models as in T2I filed \cite{FLUX}. In contrast to the image generative model community, a significant gap has emerged between open-source and closed-source video generation models. Closed-source models tend to overshadow publicly available open-source alternatives, severely limiting the potential for algorithmic innovation from the public community. While the recent state-of-the-art model MovieGen \cite{polyak2024movie} has demonstrated promising performance, its milestone for open-source release has yet to be established.

\begin{figure}[h]
    \centering
    \begin{subfigure}[b]{0.47\textwidth}
        \centering
        \includegraphics[width=\textwidth]{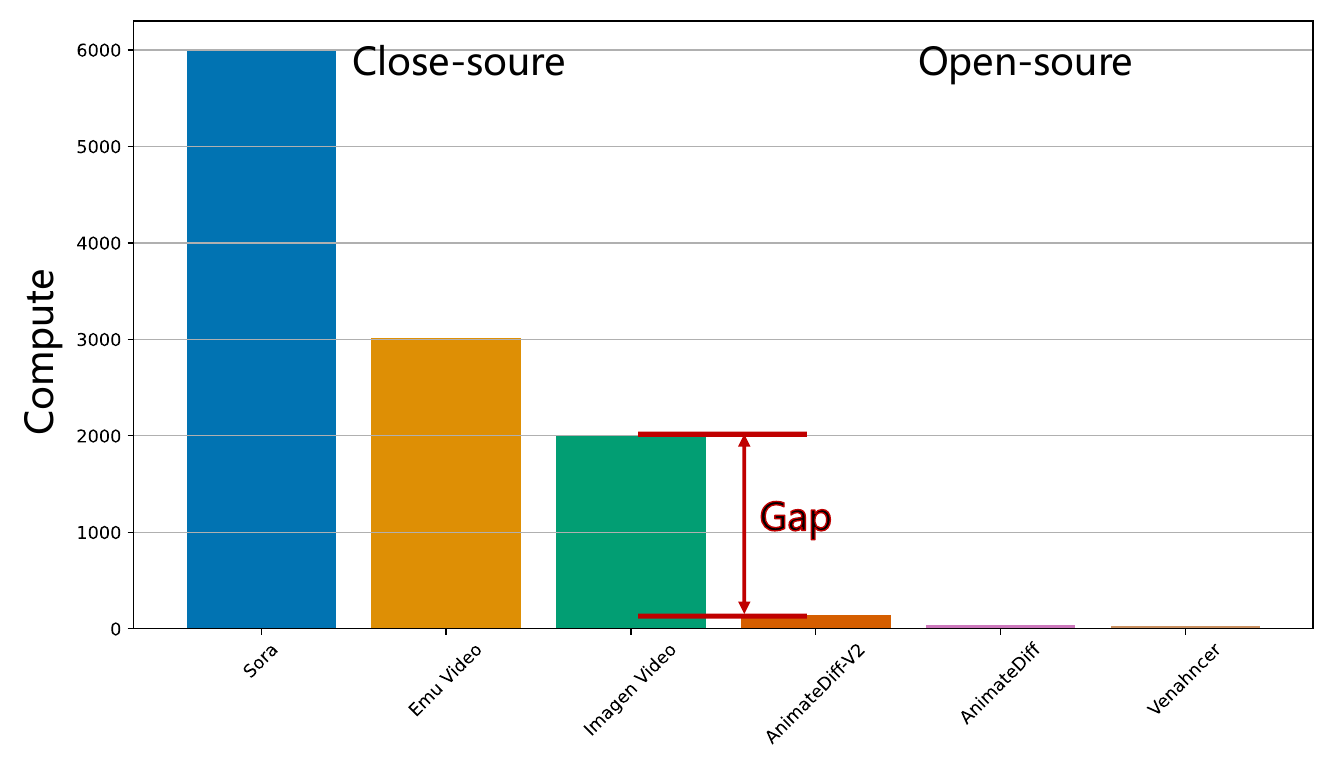} 
    \end{subfigure}
    \begin{subfigure}[b]{0.45\textwidth}
        \centering
        \ifhq
        \includegraphics[width=\textwidth]{hqfigures/ranking.pdf}
        \else
        \includegraphics[width=\textwidth]{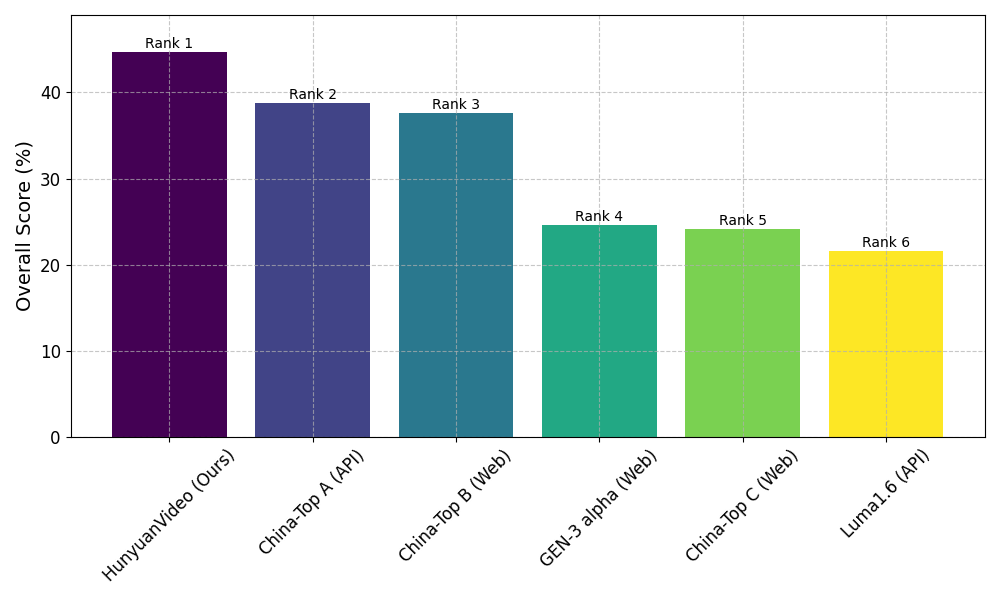}
        \fi
    \end{subfigure}
    \label{fig:side_by_side}
    \caption{Left: Computation resources used for closed-source and open-source video generation models. Right: Performance comparison between \nameofmethod{} and other selected strong baselines.}
\end{figure}

To address the existing gap and enhance the capabilities of the public community, this report presents our open-sourced foundational video generative model, \nameofmethod{}. This systematic framework encompasses training infrastructure, data curation, model architecture optimization, and model training.
Through our experiments, we discovered that randomly scaling the training data, computational resources, and model parameters of a simple Transformer-based generative model \cite{peebles2023scalable} trained with Flow Matching \cite{lipman2022flow} was not sufficiently efficient. Consequently, we explored an effective scaling strategy that can reduce computational resource requirements by up to 5× while achieving the desired model performance. With this optimal scaling approach and dedicated infrastructure, we successfully trained a large video model comprising 13 billion parameters, pre-training it on internet-scale images and videos.
After a dedicated progressive fine-tuning strategy, \nameofmethod{} excels in four critical aspects of video generation: visual quality, motion dynamics, video-text alignment, and semantic scene cut. We conducted a comprehensive comparison of \nameofmethod{} with leading global video generation models, including Gen-3 and Luma 1.6 and 3 top performing commercial models in China, using over 1,500 representative text prompts accessed by a group of 60 people. The results indicate that \nameofmethod{} achieves the highest overall satisfaction rates, particularly excelling in motion dynamics.

\section{Overview}
\label{sec:framework}
\nameofmethod{} is a comprehensive video training system encompassing all aspects from data processing to model deployment. This technical report is structured as follows:

\begin{itemize}
    \item In \textbf{Section \ref{sec:data}}, we introduce our data preprocessing techniques, including filtering and re-captioning models.
    \item \textbf{Section \ref{sec:model_arch}} presents detailed information about the architecture of all components of \nameofmethod{}, along with our training and inference strategies.
    \item In \textbf{Section \ref{sec:accelerate}}, we discuss methods for accelerating model training and inference, enabling the development of a large model with 13 billion parameters.
    \item \textbf{Section \ref{sec:exp}} evaluates the performance of our text-to-video foundation models and compares them with state-of-the-art video generation models, both open-source and proprietary.
    \item Finally, in \textbf{Section \ref{sec:application}}, we showcase various applications built on the pre-trained foundation model, accompanied by relevant visualizations as well as some video related functional models such as video to audio generative model.
\end{itemize}
\begin{figure}[h]
    \hfill
    \ifhq
    \includegraphics[width=\linewidth]{hqfigures/overall.pdf}
    \else
    \includegraphics[width=\linewidth]{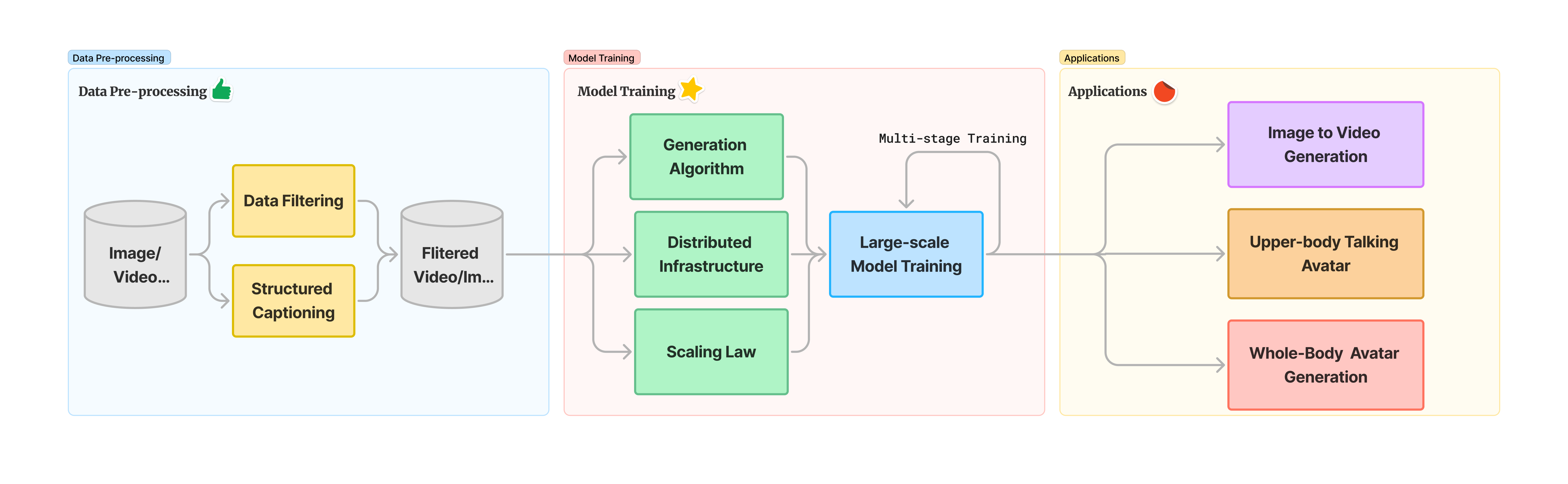}
    \fi
    \caption{The overall training system for \nameofmethod{}.}
    \label{fig:pipeline_overview}
\end{figure}

\input{tax/data}

\section{Model Architecture Design}
\begin{figure}[t]
    \centering
    \ifhq
    \includegraphics[trim={2cm 2cm 2cm 2cm},clip,width=0.95\linewidth]{hqfigures/hunyuanvideo_overview.pdf}
    \else
    \includegraphics[width=0.95\linewidth]{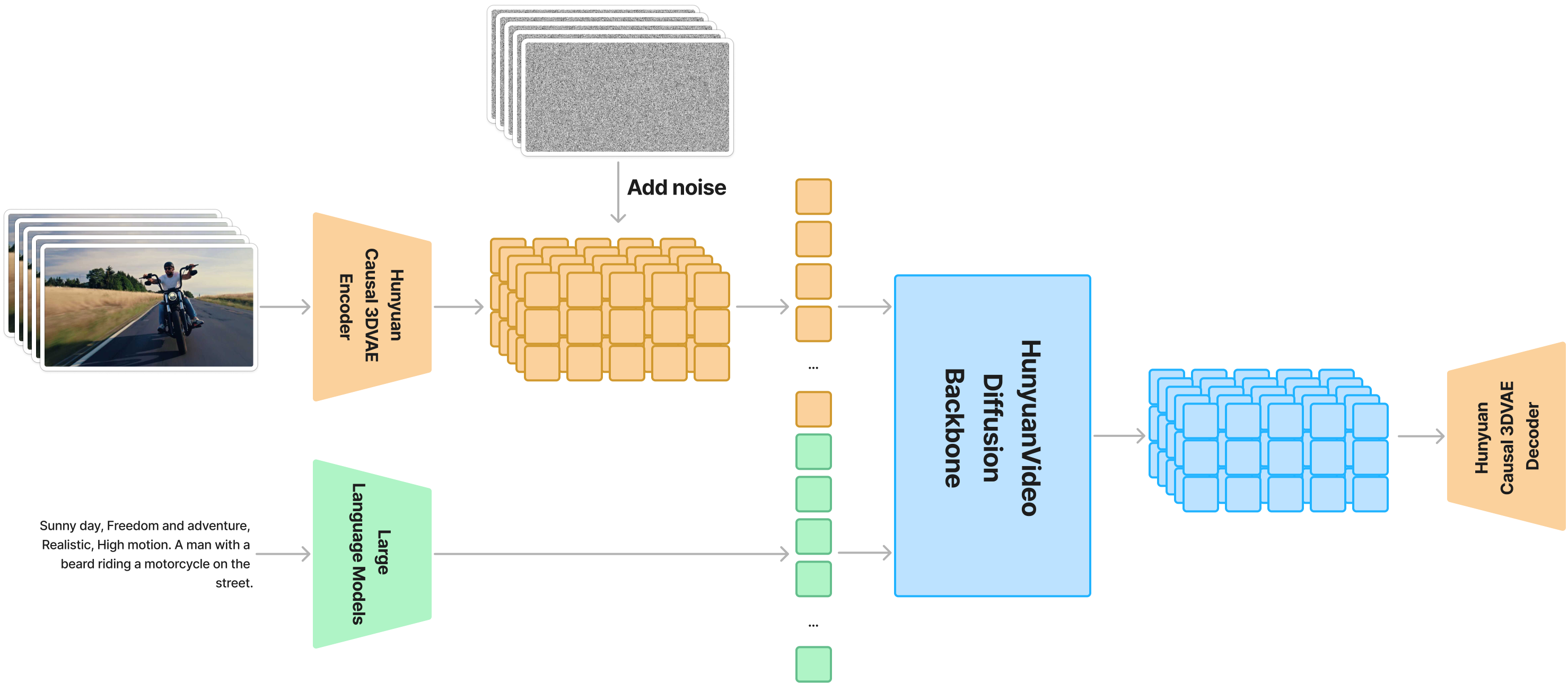}
    \fi
    \caption{The overall architecture of \nameofmethod{}. The model is trained on a spatial-temporally compressed latent space, which is compressed through Causal 3D VAE. Text prompts are encoded using a large language model, and used as the condition. Gaussian noise and condition are taken as input, our model generates a output latent, which is decoded into images or videos through the 3D VAE decoder.}
    \label{fig:hunyuanvideo_overview}
\end{figure}

The overview of our \nameofmethod{} model is shown in Fig.~\ref{fig:hunyuanvideo_overview}. This section describes the Causal 3D VAE, diffusion backbone, and scaling laws experiments.

\label{sec:model_arch}

\input{tax/vae}

\input{tax/scaling_law}

\input{tax/pre-training}


\input{tax/accelerate}

\section{Fundation Model Performance}
\label{sec:exp}

\input{exp/text_encoder}

\input{exp/sft}

\input{exp/human_eval}

\section{Applications}
\label{sec:application}

\input{tax/tv2a}

\input{applications/i2v}

\input{applications/human_i2v}

\input{applications/audio}

\input{applications/expr_pose}

\input{applications/demo}

\section{Related Works}
\label{sec:related_works}

Due to the success of diffusion models in the field of image generation~\citep{rombach2022high, ho2020denoising}, the exploration in the domain of video generation~\citep{guo2023animatediff,jiang2023text2performer,singer2022make,wang2023lavie,yang2023probabilistic,zhang2023show,ma2024follow,chen2024follow,xue2024follow,ma2024follow2} is also becoming popular. VDM~\citep{ho2022video} is among the first that extends the 2D U-Net from image diffusion models to a 3D U-Net to achieve text-based generation.
Later works, such as MagicVideo~\citep{zhou2023magicvideo} and Mindscope~\citep{wang2023modelscope}, introduce 1D temporal attention mechanisms, reducing computations by building upon latent diffusion models. In this report, we do not use the 2D + 1D temporal block manner for motion learning. Instead, we use similar dual flow attention blocks as in FLUX \cite{FLUX}, which are used for processing all video frames.
Following Imagen, Imagen Video~\citep{ho2022imagen} employs a cascaded sampling pipeline that generates videos through multiple stages.
In addition to traditional end-to-end text-to-video (T2V) generation, 
video generation using other conditions is also an important direction.
This type of methods generates videos with other auxiliary controls, such as depth maps~\citep{guo2023sparsectrl,he2023animate}, pose maps~\citep{xu2023magicanimate,hu2023animate,wang2023disco,ma2023follow}, RGB images~\citep{blattmann2023stable,chen2023seine,ni2023conditional}, or other guided motion videos~\citep{zhao2023motiondirector,wu2023lamp}.  
Despite the excellent generation performance of the recent open-source models such as Stable video diffusion~\citep{blattmann2023stable}, Open-sora \cite{opensora}, Open-sora-plan \cite{pku_yuan_lab_and_tuzhan_ai_etc_2024_10948109}, Mochi-1 \cite{genmo2024mochi} and Allegro \cite{zhou2024allegro}, their performance still falls far behind the closed-source
state-of-the-art video generation models such as Sora \cite{videoworldsimulators2024} and MovieGen \cite{polyak2024movie}.

\input{tax/contributors}

\clearpage
{
\bibliographystyle{plain}
\bibliography{egbib}
}

\end{document}

%% file: tax/data.tex
\section{Data Pre-processing}
\label{sec:data}


We use an image-video joint training strategy.
The videos are meticulously divided into five distinct groups, while images are categorized into two groups, each tailored to fit the specific requirements of their respective training processes. This section will primarily delve into the intricacies of video data curation.

Our data acquisition process is rigorously governed by the principles outlined in the General Data Protection Regulation (GDPR) \cite{investopedia_gdpr} framework. Furthermore, we employ advanced techniques such as data synthesis and privacy computing to guarantee compliance with these stringent standards.

Our raw data pool initially comprised videos spanning a wide range of domains including people, animals, plants, landscapes, vehicles, objects, buildings, and animation. Each video was acquired with a set of basic thresholds, including minimum duration requirements. Additionally, a subset of the data was collected based on more stringent criteria, such as spatial quality, adherence to a specific aspect ratio, and professional standards in composition, color, and exposure. These rigorous standards ensure that our videos possess technical quality and aesthetic appeal. We experimentally verified that incorporating high-quality data is instrumental in significantly enhancing model performance.

\begin{figure}[t]
    \centering
    \includegraphics[trim=1cm 2cm 2cm 1cm,width=\linewidth]{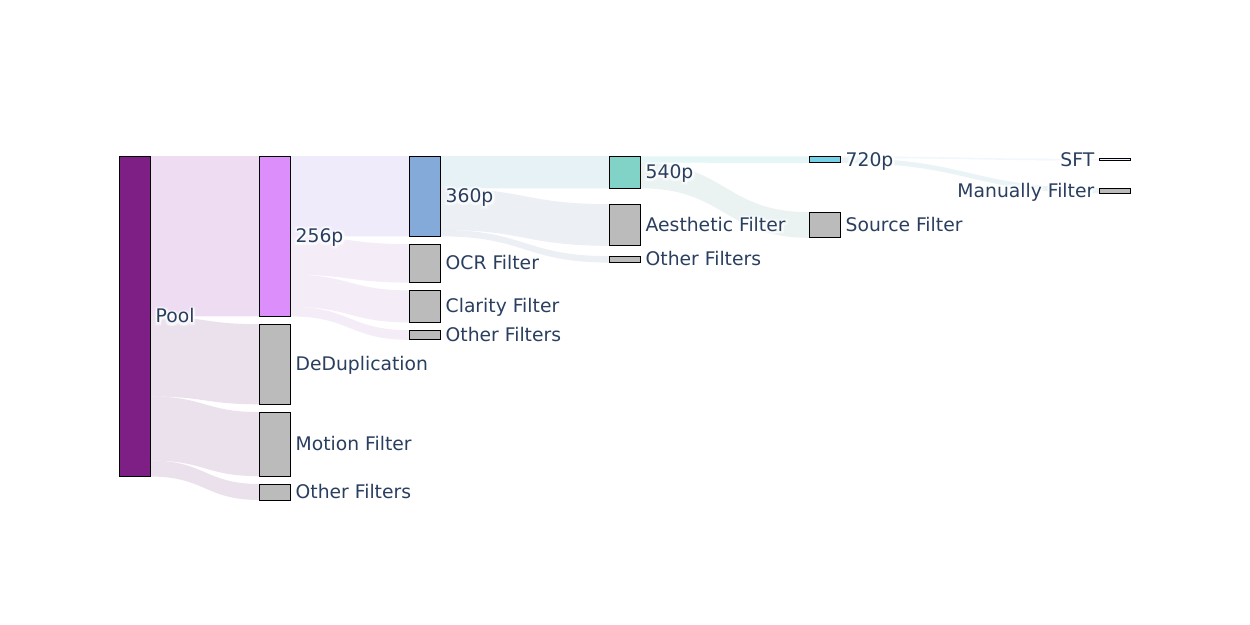}
     \caption{
     \textbf{Our hierarchical data filtering pipeline.} 
     We employ various filters for data filtering and progressively increase their thresholds to build 4 training datasets, i.e., 256p, 360p, 540p, and 720p, while the final SFT dataset is built through manual annotation. This figure highlights some of the most important filters to use at each stage. A large portion of data will be removed at each stage, ranging from half to one-fifth of the data from the previous stage. 
     Here, gray bars represent the amount of data filtered out by each filter while colored bars indicate the amount of remaining data at each stage.
     }
    \label{fig:video-data-curation}
\end{figure}

\subsection{Data Filtering}
\label{data_filtering}

Our raw data from different sources exhibits varying durations and levels of quality. To address this, we employ a series of techniques to pre-process the raw data. Firstly, we utilize PySceneDetect~\cite{pyscene} to split raw videos into single-shot video clips. Next, we employ the Laplacian operator from OpenCV~\cite{opencv} to identify a clear frame, serving as the starting frame of each video clip. Using an internal VideoCLIP model, we calculate embeddings for these video clips. These embeddings serve two purposes: (i) we deduplicate similar clips based on the Cosine distance of their embeddings; (ii) we apply k-means~\cite{macqueen1967some} to obtain $\sim$10K concept centroids for concept resampling and balancing.

To continuously enhance video aesthetics, motion, and concept range, we implement a hierarchical data filtering pipeline for constructing training datasets, as shown in Figure~\ref{fig:video-data-curation}. This pipeline incorporates various \textit{filters} to help us filter data from different perspectives which we introduce next. 

We employ Dover~\cite{wu2023exploring} to assess the visual aesthetics of video clips from both aesthetic and technical viewpoints. Additionally, we train a model to determine clarity and eliminate video clips with visual blurs. By predicting the motion speed of videos using estimated optical flow~\cite{opencv}, we filter out static or slow-motion videos. We combine the results from PySceneDetect~\cite{pyscene} and Transnet v2~\cite{souvcek2020transnet} to get scene boundary information. We utilize an internal OCR model to remove video clips with excessive text, as well as to locate and crop subtitles. We also develop YOLOX~\cite{ge2021yolox}-like visual models to detect and remove some occluded or sensitive information such as watermarks, borders, and logos. To assess the effectiveness of these filters, we perform simple experiments using a smaller \nameofmethod{} model and observe the performance changes. The results obtained from these experiments play an important role in guiding the building of our data filtering pipeline, which is introduced next.

Our hierarchical data filtering pipeline for video data yields five training datasets, corresponding to the five training stages (Section ~\ref{model-pretrain}). These datasets (except for the last fine-tuning dataset) are curated by progressively improving the thresholds of the aforementioned filters. The video spatial resolution increases progressively from 256 $\times$ 256 $\times$ 65 to 720$\times$1280 $\times$ 129.
%
%
We apply varying levels of strictness to the filters during the threshold adjustment process at different stages (see Figure~\ref{fig:video-data-curation}). The last dataset used for fine-tuning is described next. 

To improve the model's performance in the final stage (Section~\ref{sft}), we build a fine-tuning dataset comprising $\sim$1M samples. This dataset is meticulously curated through human annotation. Annotators are assigned the task of identifying video clips that exhibit high visual aesthetics and compelling content motion. Each video clip undergoes evaluation based on two perspectives: (i) decomposed aesthetical views, including color harmony, lighting, object emphasis, and spatial layout; (ii) decomposed motion views, encompassing motion speed, action integrity, and motion blurs. Finally, our fine-tuning dataset consists of visually appealing video clips with intricate motion details.

We also establish a hierarchical data filtering pipeline for images by reusing most of the filters, excluding the motion-related ones. Similarly, we build two image training datasets by progressively increasing the filtering thresholds applied to an image pool of billions of image-text pairs. 
The first dataset contains billions of samples and is used for the initial stage of text-to-image pre-training. The second dataset contains hundreds of millions of samples and is utilized for the second stage of text-to-image pre-training.

\subsection{Data Annotation}

\textbf{Structured Captioning}. As evidenced in research \cite{videoworldsimulators2024,betker2023improving}, the precision and comprehensiveness of captions play a crucial role in improving the prompt following ability and output quality of generative models. Most previous work focus on providing either brief captions \cite{Chen_2024,li2022blip} or dense captions \cite{yang2024cogvideox,chen2023sharegpt4v,chen2024sharegpt4video}. However, these approaches are not without shortcomings, suffering from incomplete information, redundant discourse and inaccuracies. In pursuit of achieving captions with higher comprehensiveness, information density and accuracy, we develop and implement an in-house Vision Language Model(VLM) designed to generate structured captions for images and videos. These structured captions, formatted in JSON, provide multi-dimensional descriptive information from various perspectives, including: 1) \textbf{Short Description}: Capturing the main content of the scene. 2) \textbf{Dense Description}: Detailing the scene's content, which notably includes scene transitions and camera movements that are integrated with the visual content, such as camera follows some subject. 3) \textbf{Background}: Describing the environment in which the subject is situated. 4) \textbf{Style}: Characterizing the style of the video, such as documentary, cinematic, realistic, or sci-fi. 5) \textbf{Shot Type}: Identifying the type of video shot that highlights or emphasizes specific visual content, such as aerial shot, close-up shot, medium shot, or long shot. 6) \textbf{Lighting}: Describing the lighting conditions of the video. 7) \textbf{Atmosphere}: Conveying the atmosphere of the video, such as cozy, tense, or mysterious.

Moreover, we extend the JSON structure to  incorporate additional metadata-derived elements, including source tags, quality tags, and other pertinent tags from meta information of images and videos. Through the implementation of a carefully designed dropout mechanism coupled with permutation and combination strategies, we synthesize captions diverse in length and pattern by assembling these multi-dimensional descriptions for each image and video, aiming to improve the generalization ability of generative models and prevent overfitting. We utilize this captioner to provide structured captions for all images and videos in our training dataset.

\textbf{Camera Movement Types}. We also train a camera movement classifier capable of predicting 14 distinct camera movement types, including zoom in, zoom out, pan up, pan down, pan left, pan right, tilt up, tilt down, tilt left, tilt right, around left, around right, static shot and handheld shot. High-confidence predictions of camera movement types are integrated into the JSON-formatted structured captions, to enable camera movement control ability of generative models.


%% file: tax/vae.tex
\subsection{3D Variational Auto-encoder Design}\label{3dVAE}

\begin{figure}[t]
    \centering
    \includegraphics[width=0.95\linewidth]{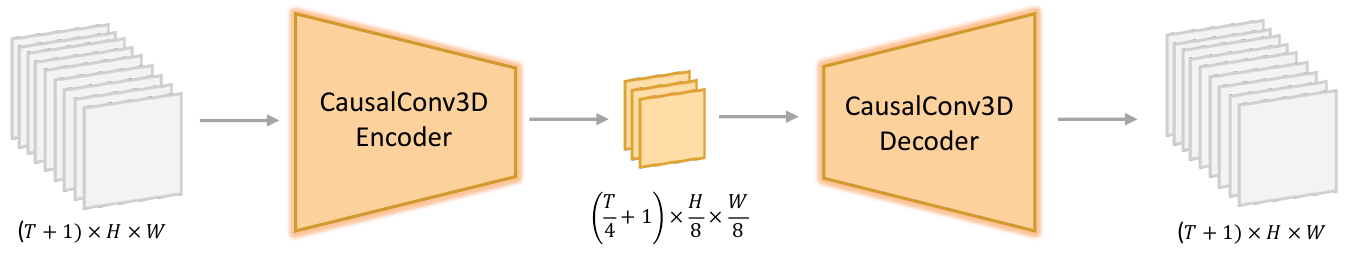}
    \caption{ The architecture of our 3DVAE.}
    \label{fig:vae-model-arch}
\end{figure}

Similar to previous work~\cite{polyak2024movie,yang2024cogvideox}, we train a 3DVAE to compress pixel-space videos and images into a compact latent space. To handle both videos and images, we adopt CausalConv3D~\cite{yu2023language}. For a video of shape $(T+1) \times 3 \times H \times W$, our 3DVAE compresses it into latent features with shape $(\frac{T}{c_t} + 1) \times C \times (\frac{H}{c_s}) \times (\frac{W}{c_s})$. In our implementation, $c_t=4$, $c_s=8$, and $C=16$. This compression significantly reduces the number of tokens for the subsequent diffusion transformer model, allowing us to train videos at the original resolution and frame rate. The model structure is illustrated in Figure \ref{fig:vae-model-arch}.

\subsubsection{Training}
In contrast to most previous work \cite{polyak2024movie,chen2024od,zhou2024allegro}, we do not rely on a pre-trained image VAE for parameter initialization; instead, we train our model from scratch. 
To balance the reconstruction quality of videos and images, we mix video and image data at a ratio of $4:1$. Besides the routinely used $L_1$ reconstruction loss and KL loss $L_{kl}$, we also incorporate perceptual loss $L_{lpips}$ and GAN adversarial loss $L_{adv}$ \cite{esser2021taming} to enhance the reconstruction quality. The complete loss function is shown in Equation \ref{eq:vae-loss}.

\begin{equation}
    \label{eq:vae-loss}
    \text{Loss} = L_{1} + 0.1 L_{lpips} + 0.05 L_{adv} + 10^{-6} L_{kl}
\end{equation}

During training, we employ a curriculum learning strategy, gradually training from low-resolution short video to high-resolution long video. To improve the reconstruction of high-motion videos, we randomly choose a sampling interval from the range $1 \sim 8$ to sample frames evenly across video clips.

\subsubsection{Inference}
Encoding and decoding high-resolution long videos on a single GPU can lead to out-of-memory (OOM) errors. To address this, we use a spatial-temporal tiling strategy, splitting the input video into overlapping tiles along the spatial and temporal dimensions. Each tile is encoded/decoded separately, and the outputs are stitched together. For the overlapping regions, we utilize a linear combination for blending. This tiling strategy allows us to encode/decode videos in arbitrary resolutions and durations on a single GPU.

We observed that directly using the tiling strategy during inference can result in visible artifacts due to inconsistencies between training and inference. To solve this, we introduce an additional finetuning phase where the tiling strategy is randomly enabled/disabled during training. This ensures the model is compatible with both tiling and non-tiling strategies, maintaining consistency between training and inference. 

\begin{figure}[ht]
    \centering
    \ifhq
    \includegraphics[width=\linewidth]{hqfigures/vae-sota-cmp.png}
    \else
    \includegraphics[width=\linewidth]{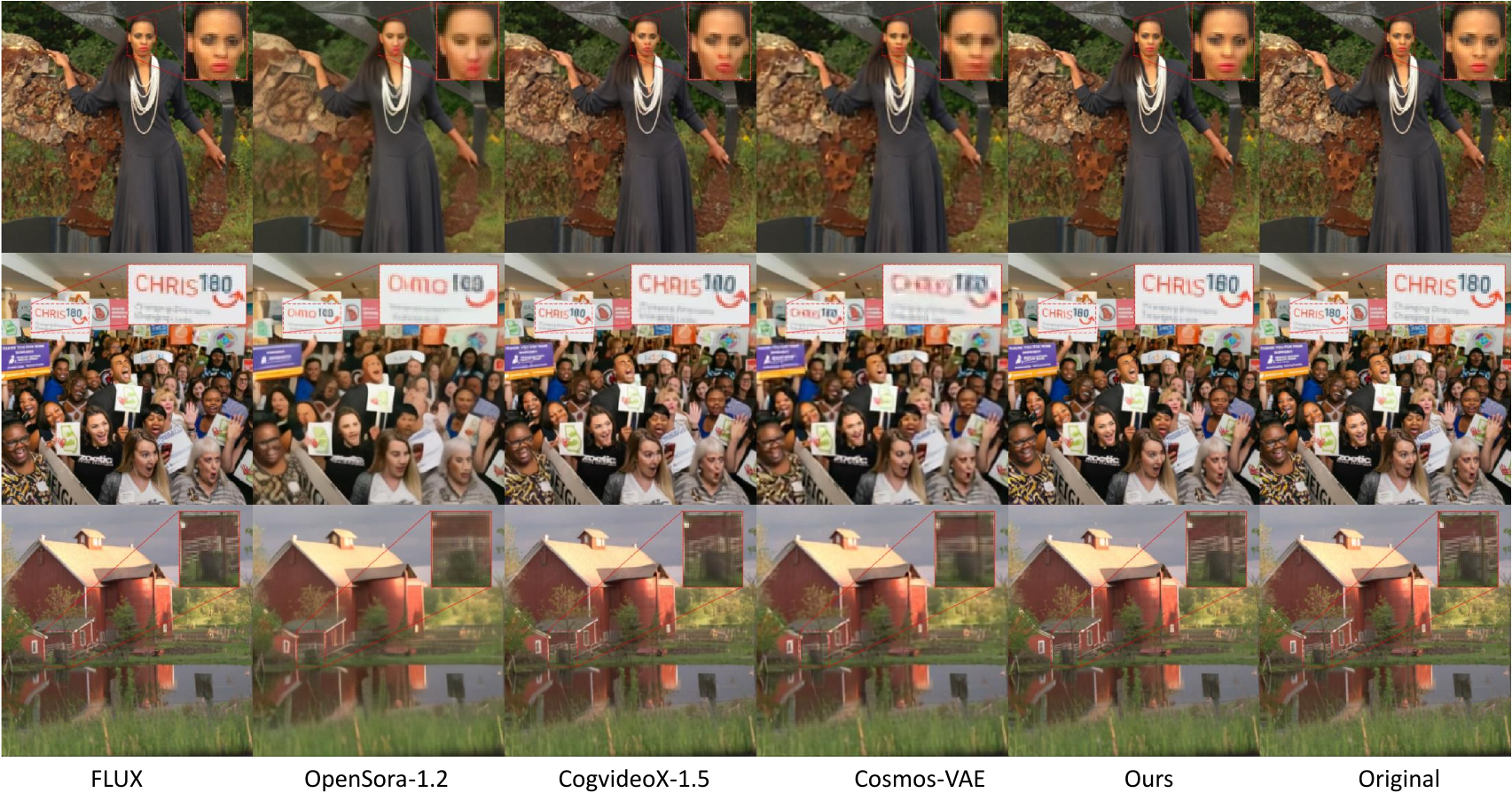}
    \fi
    \caption{VAE reconstruction case comparison.}
    \label{fig:vae-sota-cmp}
\end{figure}

Table \ref{tab:sota_vae} compares our VAE with open-source state-of-the-art VAEs. On video data, our VAE demonstrates a significantly higher PSNR compared to other video VAEs. On images, our performance surpasses both video VAEs and image VAE. Figure \ref{fig:vae-sota-cmp} shows several cases at $256 \times 256$ resolution. Our VAE demonstrates significant advantages in text, small faces, and complex textures.

\begin{table*}[ht]
\renewcommand{\arraystretch}{1.2}
\small
\centering  
\caption{VAE reconstruction metrics comparison.}
\begin{tabular}{lcccc}
\toprule
\multirow{2}{*}{Model} & Downsample  & \multirow{2}{*}{$|z|$} & ImageNet (256$\times$256)
& MCL-JCV (33$\times$360$\times$640) \\
& Factor & & PSNR$\uparrow$ & PSNR$\uparrow$ \\
\midrule
FLUX-VAE~\cite{FLUX}                   & $1 \times 8 \times 8$ & 16 & 32.70 & - \\
\midrule
OpenSora-1.2~\cite{opensora}           & $4 \times 8 \times 8$ & 4  & 28.11 & 30.15 \\
CogvideoX-1.5~\cite{yang2024cogvideox} & $4 \times 8 \times 8$ & 16 & 31.73 & 33.22 \\
Cosmos-VAE~\cite{cosmos}               & $4 \times 8 \times 8$ & 16 & 30.07 & 32.76 \\
Ours                                   & $4 \times 8 \times 8$ & 16 & 33.14 & 35.39 \\
\bottomrule
\end{tabular}
\label{tab:sota_vae}
\end{table*}

%% file: tax/scaling_law.tex
\subsection{Unified Image and Video Generative Architecture}
\label{sec:architecture}


\begin{figure}[ht]
    \centering
    \includegraphics[width=\linewidth]{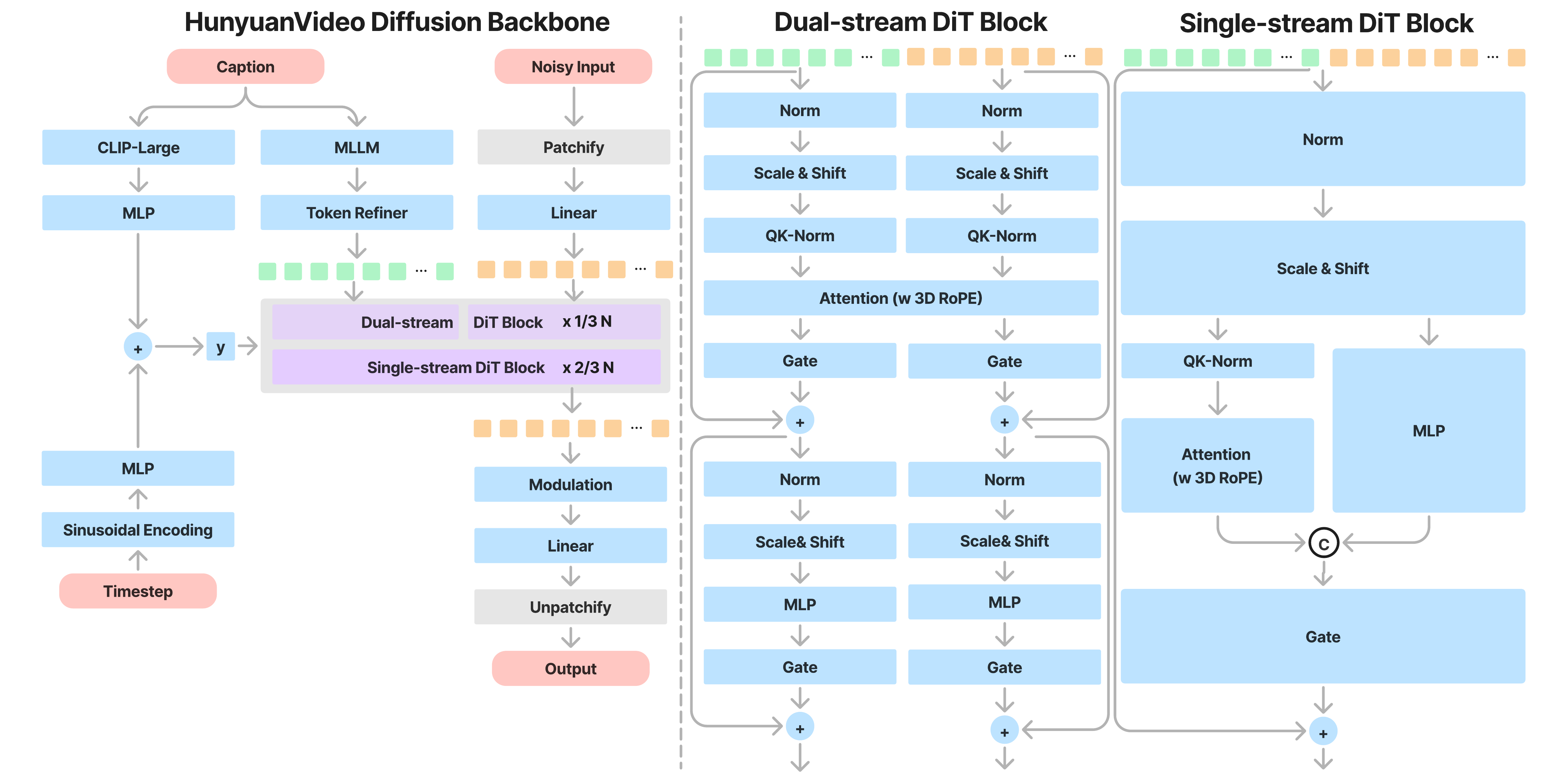}
    \caption{ The architecture of our \nameofmethod{} Diffusion Backbone.}
    \label{fig:dit_bnackbone}
\end{figure}

In this section, we introduce the Transformer design in \nameofmethod{}, which employs a unified Full Attention mechanism for three main reasons:
Firstly, it has demonstrated superior performance compared to divided spatiotemporal attention~\cite{videoworldsimulators2024,polyak2024movie,yang2024cogvideox,genmo2024mochi}.
Secondly, it supports unified generation for both images and videos, simplifying the training process and improving model scalability.
Lastly, it leverages existing LLM-related acceleration capabilities more effectively, enhancing both training and inference efficiency.
The model structure is illustrated in Figure \ref{fig:dit_bnackbone}.

\textbf{Inputs.} For a given video-text pair, the model operates within the 3D latent space described in Section \ref{3dVAE}. Specifically, for the video branch, the input is first compressed into latents of shape $T \times C \times H \times W$. To unify input processing, we treat images as single-frame videos. These latents are then patchified and unfolded into a 1D sequence of tokens with a length of $\frac{T}{k_t} \cdot \frac{H}{k_h} \cdot \frac{W}{k_w}$ using a 3D convolution with a kernel size of $k_t \times k_h \times k_w$. For the text branch, we first use an advanced LLM to encode the text into a sequence of embeddings that capture fine-grained semantic information. Concurrently, we employ the CLIP model to extract a pooled text representation containing global information. This representation is then expanded in dimensionality and added to the timestep embedding before being fed into the model.

\textbf{Model Design.}  To integrate textual and visual information effectively, we follow a similar strategy of "Dual-stream to Single-stream" hybrid model design as introduced in \cite{FLUX} for video generation. In the dual-stream phase, video and text tokens are processed independently through multiple Transformer blocks, enabling each modality to learn its own appropriate modulation mechanisms without interference. In the single-stream phase, we concatenate the video and text tokens and feed them into subsequent Transformer blocks for effective multimodal information fusion. This design captures complex interactions between visual and semantic information, enhancing overall model performance.

\textbf{Position Embedding.} To support multi-resolution, multi-aspect ratio, and varying duration generation, we use Rotary Position Embedding (RoPE)~\cite{su2023roformer} in each Transformer block. RoPE applies a rotary frequency matrix to the embeddings, enhancing the model's ability to capture both absolute and relative positional relationships, and demonstrating some extrapolation capability in LLMs. Given the added complexity of the temporal dimension in video data, we extend RoPE to three dimensions. Specifically, we compute the rotary frequency matrix separately for the coordinates of time ($T$), height ($H$), and width ($W$). We then partition the feature channels of the query and key into three segments $(d_t, d_h, d_w)$, multiply each segment by the corresponding coordinate frequencies and concatenate the segments. This process yields position-aware query and key embeddings, which are used for attention computation.


For detailed model settings, please refer to Table~\ref{tab:model_settings}.
\begin{table}[t]
  \centering
  \footnotesize
  \caption{Architecture hyperparameters for the \nameofmethod{} 13B parameter foundation model.}
  \begin{tabular}{ccccccc}
  \toprule
  \textbf{\makecell{Dual-stream \\Blocks}} & \textbf{\makecell{Single-stream \\Blocks}} & \textbf{\makecell{Model \\Dimension}} & \textbf{\makecell{FFN \\Dimension}} & \textbf{\makecell{Attention \\Heads}} & \textbf{\makecell{Head dim}} & $(d_t, d_h, d_w)$	\\
  \midrule
  20 & 40 & 3072 & 12288 & 24 & 128 & (16, 56, 56) \\
  \bottomrule
  \end{tabular}%
  \label{tab:model_settings}
\end{table}


\subsection{Text encoder}

\begin{figure}[t]
    \centering
    \includegraphics[width=0.9\linewidth]{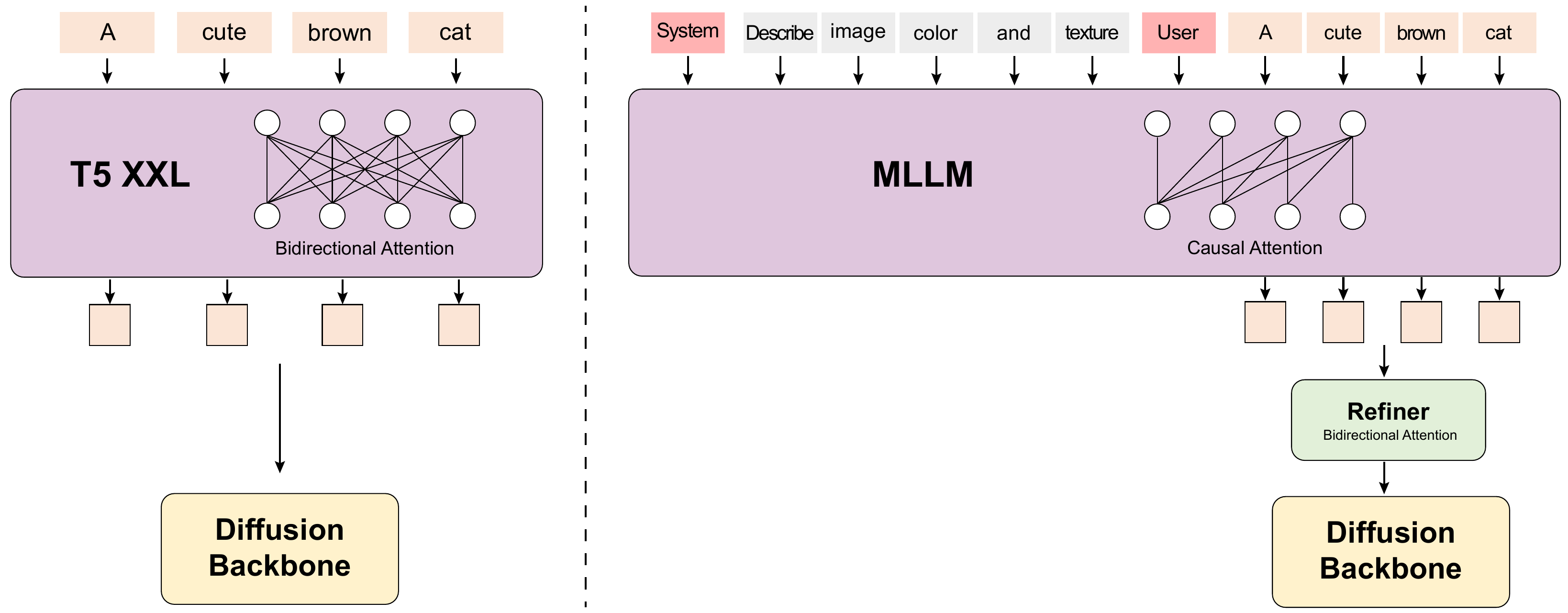}
    \caption{Text encoder comparison between T5 XXL and the instruction-guided MLLM introduced by \nameofmethod{}.}
    \label{fig:text-encoder}
\end{figure}

In generation tasks like text-to-image and text-to-video, the text encoder plays a crucial role by providing guidance information in the latent space. Some representative works~\cite{podell2023sdxl, esser2024scaling, li2024hunyuandit} typically use pre-trained CLIP~\cite{radford2021learning} and T5-XXL~\cite{raffel2020exploring} as text encoders where CLIP uses Transformer Encoder and T5 uses an Encoder-Decoder structure. In contrast, we utilize a pre-trained Multimodal Large Language Model~(MLLM) with a Decoder-Only structure as our text encoder, which has following advantages: (i) Compared with T5, MLLM after visual instruction finetuning has better image-text alignment in the feature space, which alleviates the difficulty of instruction following in diffusion models; (ii) Compared with CLIP, MLLM has been demonstrated superior ability in image detail description and complex reasoning~\cite{liu2024visual}; (iii) MLLM can play as a zero-shot learner~\cite{brown2020language} by following system instructions prepended to user prompts, helping text features pay more attention to key information. In addition, as shown in Fig.~\ref{fig:text-encoder}, MLLM is based on causal attention while T5-XXL utilizes bidirectional attention that produces better text guidance for diffusion models. Therefore, we follow~\cite{ma2024exploring} to introduce an extra bidirectional token refiner for enhancing text features.
We have configured \nameofmethod{} with a series of MLLMs~\cite{sun2024hunyuanlargeopensourcemoemodel,2023xtuner,glm2024chatglm} for different purposes. Under each setting, MLLMs have shown superior performance over conventional text encoder.

In addition, CLIP text features are also valuable as the summary of the text information. As shown in Fig.~\ref{fig:dit_bnackbone} We adopt the final non-padded token of CLIP-Large text features as a global guidance, integrating into the dual-stream and single-stream DiT blocks. 

\subsection{Model Scaling}
Neural scaling laws~\cite{kaplan2020scaling,hoffmann2022training} in language model training offer a powerful tool for understanding and optimizing the performance of machine learning models. By elucidating the relationships between model size ($N$), dataset size ($D$), and computational resources ($C$), these laws help drive the development of more effective and efficient models, ultimately advancing the success of large model training.

\begin{figure}[t]
    \centering
    \begin{subfigure}{0.325\textwidth}
        \centering
        \includegraphics[width=\textwidth]{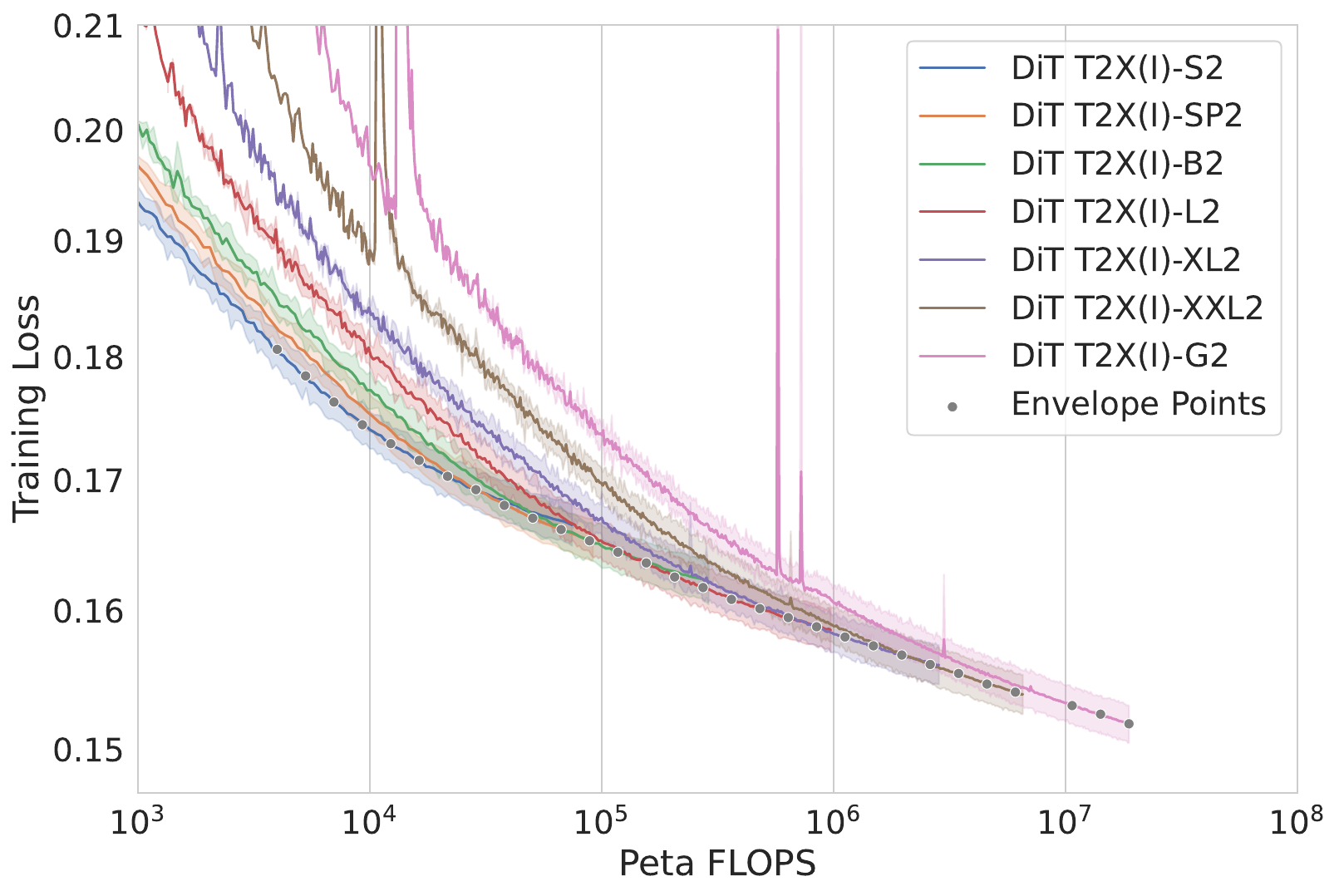}
        \caption{T2X(I) Loss curve and envelope}
        \label{fig:scaling-laws}
    \end{subfigure}
    \hfill
    \begin{subfigure}{0.325\textwidth}
        \centering
        \includegraphics[width=\textwidth]{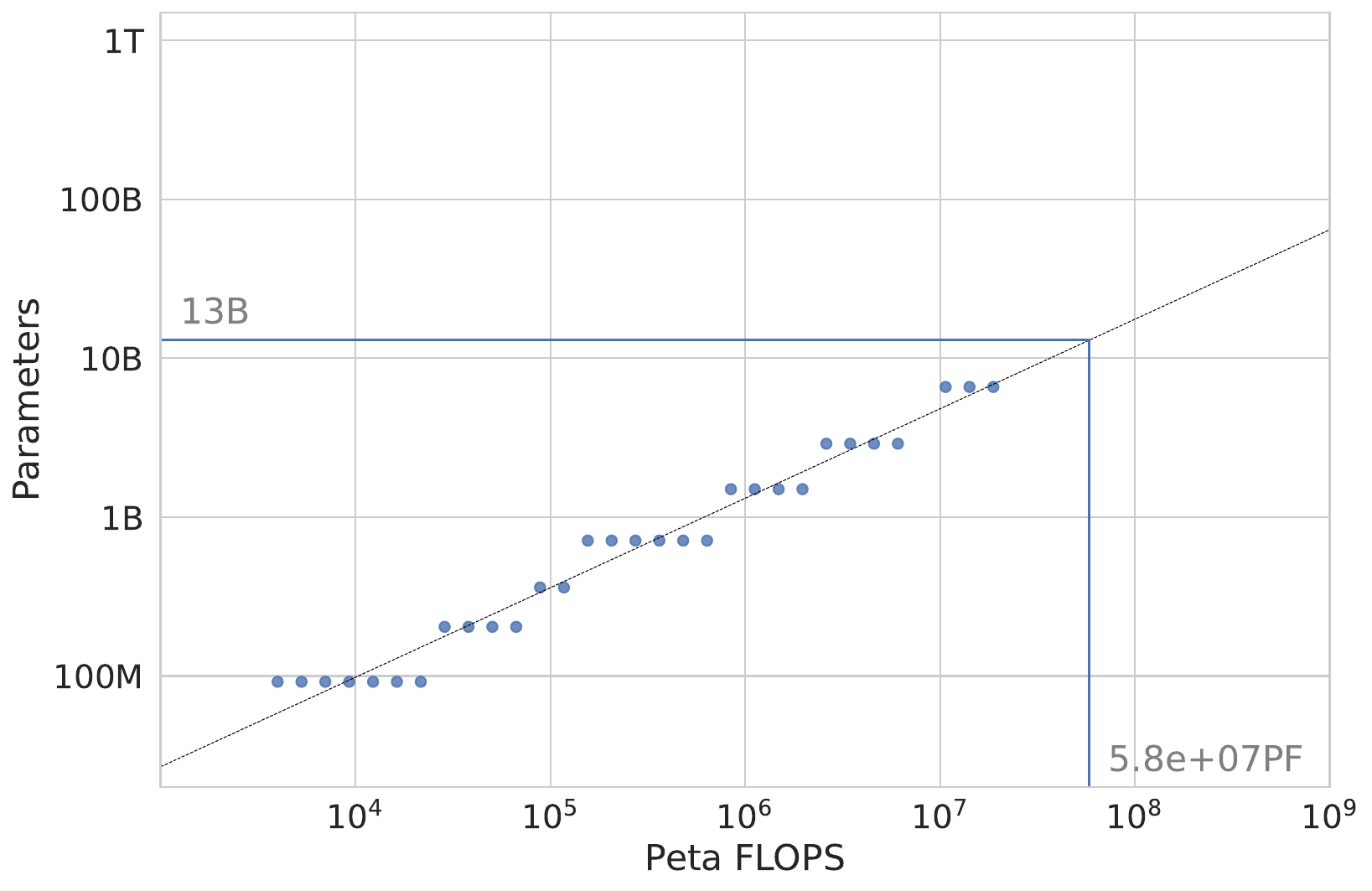}
        \caption{T2X(I) Power law of $C$ and $N$}
        \label{fig:computation-vs-parameter}
    \end{subfigure}
    \hfill
    \begin{subfigure}{0.325\textwidth}
        \centering
        \includegraphics[width=\textwidth]{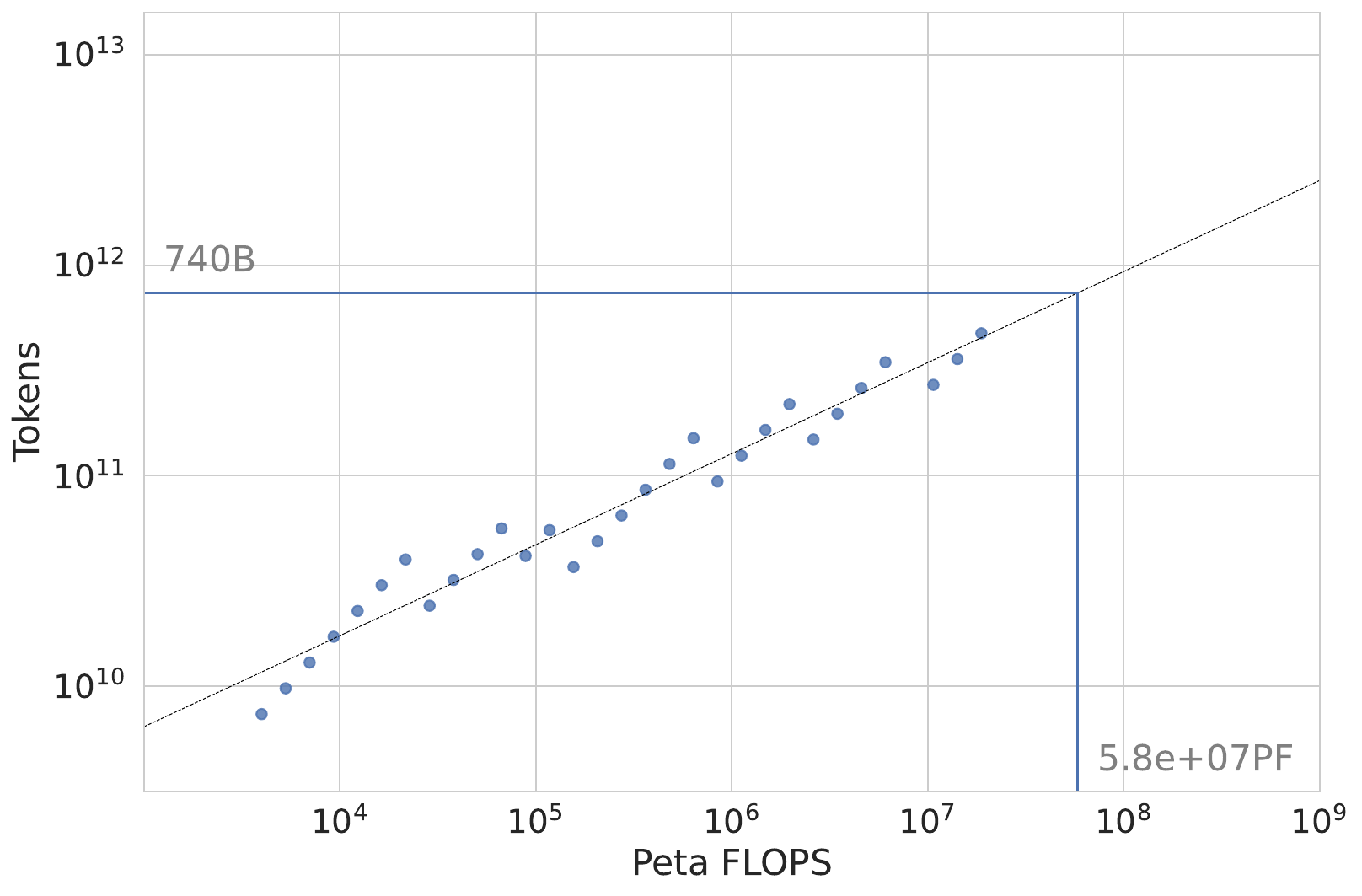}
        \caption{T2X(I) Power law of $C$ and $D$}
        \label{fig:computation-vs-token}
    \end{subfigure}
    \hfill
    \begin{subfigure}{0.325\textwidth}
        \centering
        \includegraphics[width=\textwidth]{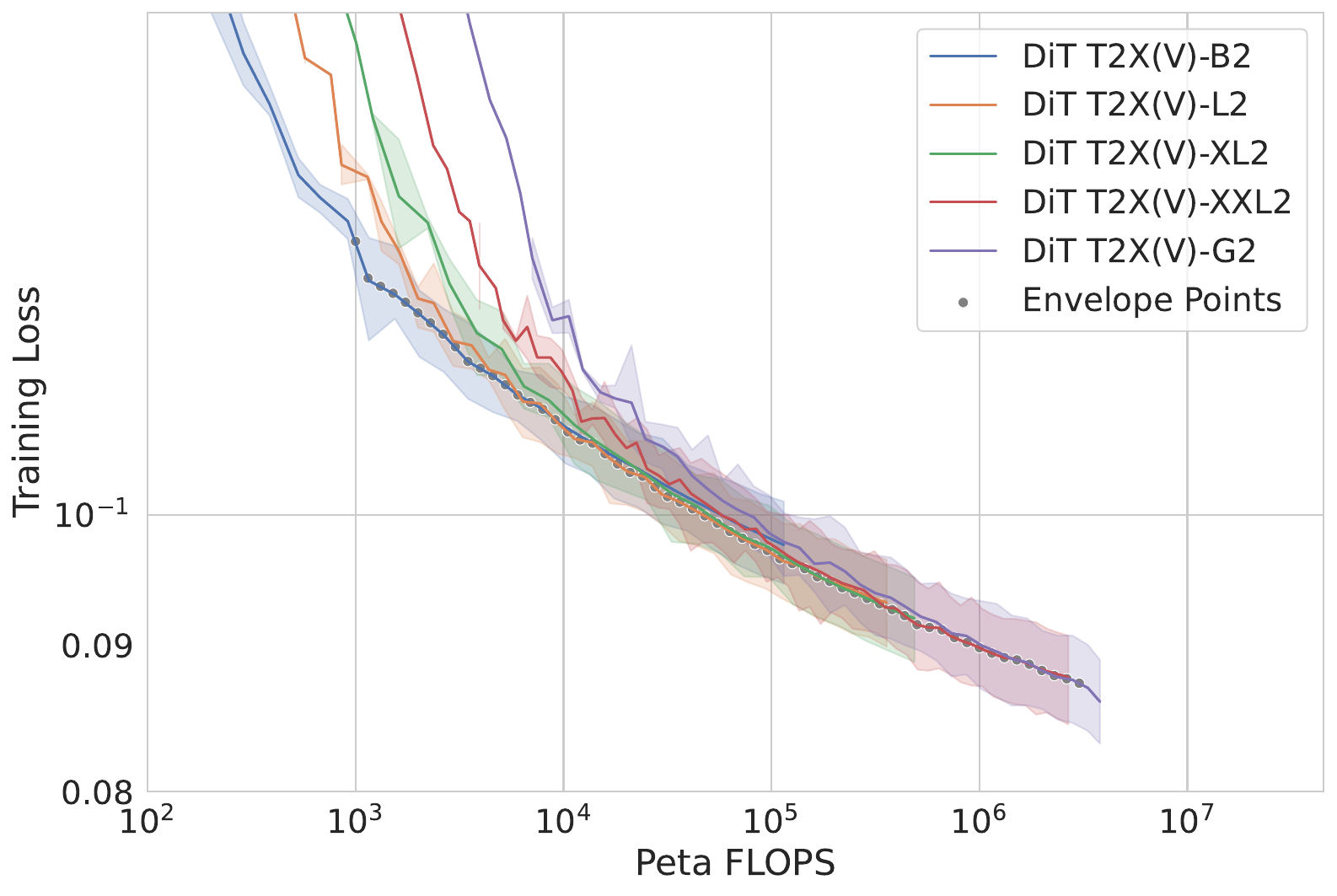}
        \caption{T2X(V) Loss curve and envelope}
        \label{fig:video-scaling-laws}
    \end{subfigure}
    \hfill
    \begin{subfigure}{0.325\textwidth}
        \centering
        \includegraphics[width=\textwidth]{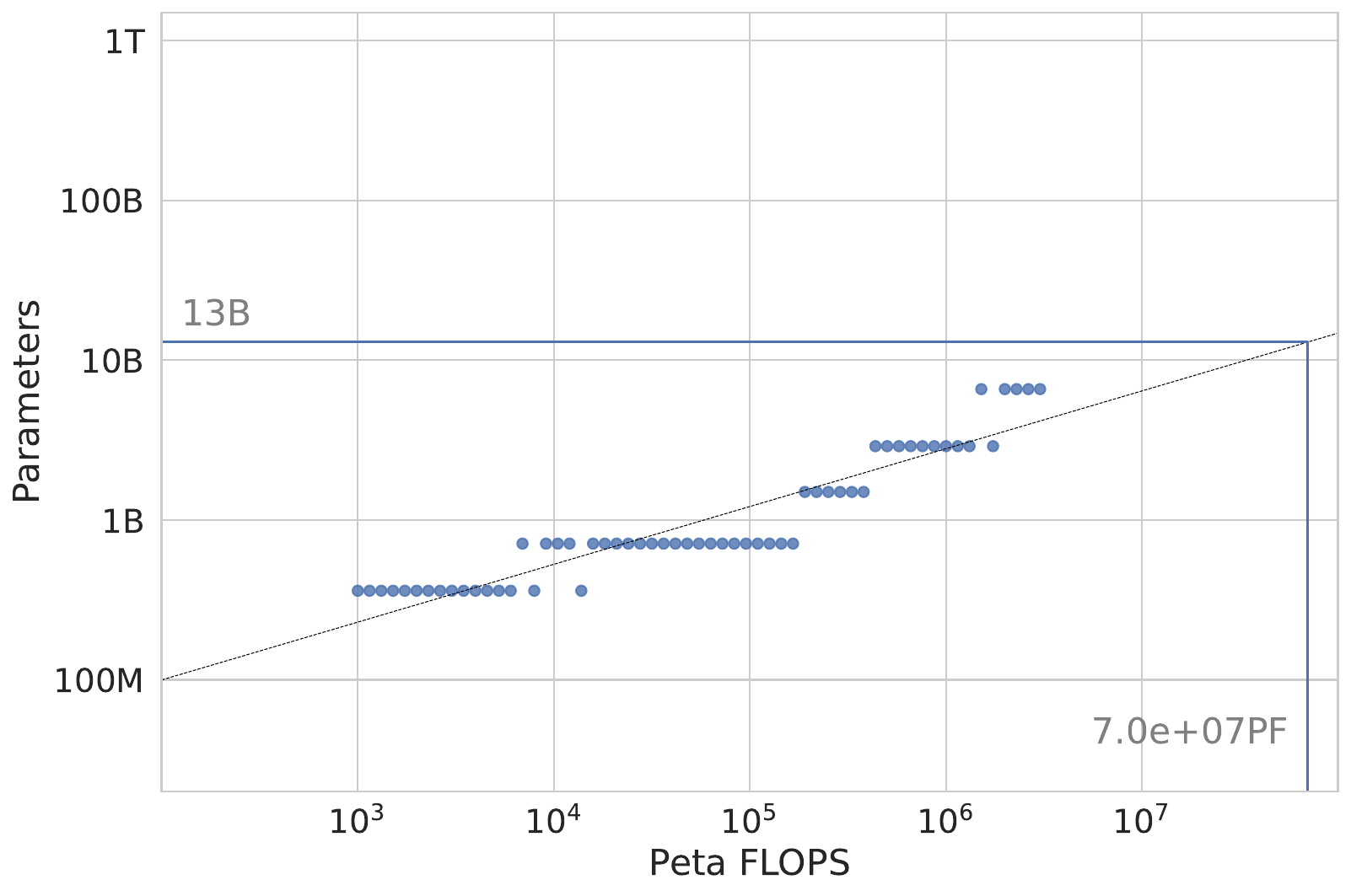}
        \caption{T2X(V) Power law of $C$ and $N$}
        \label{fig:video-computation-vs-parameter}
    \end{subfigure}
    \hfill
    \begin{subfigure}{0.325\textwidth}
        \centering
        \includegraphics[width=\textwidth]{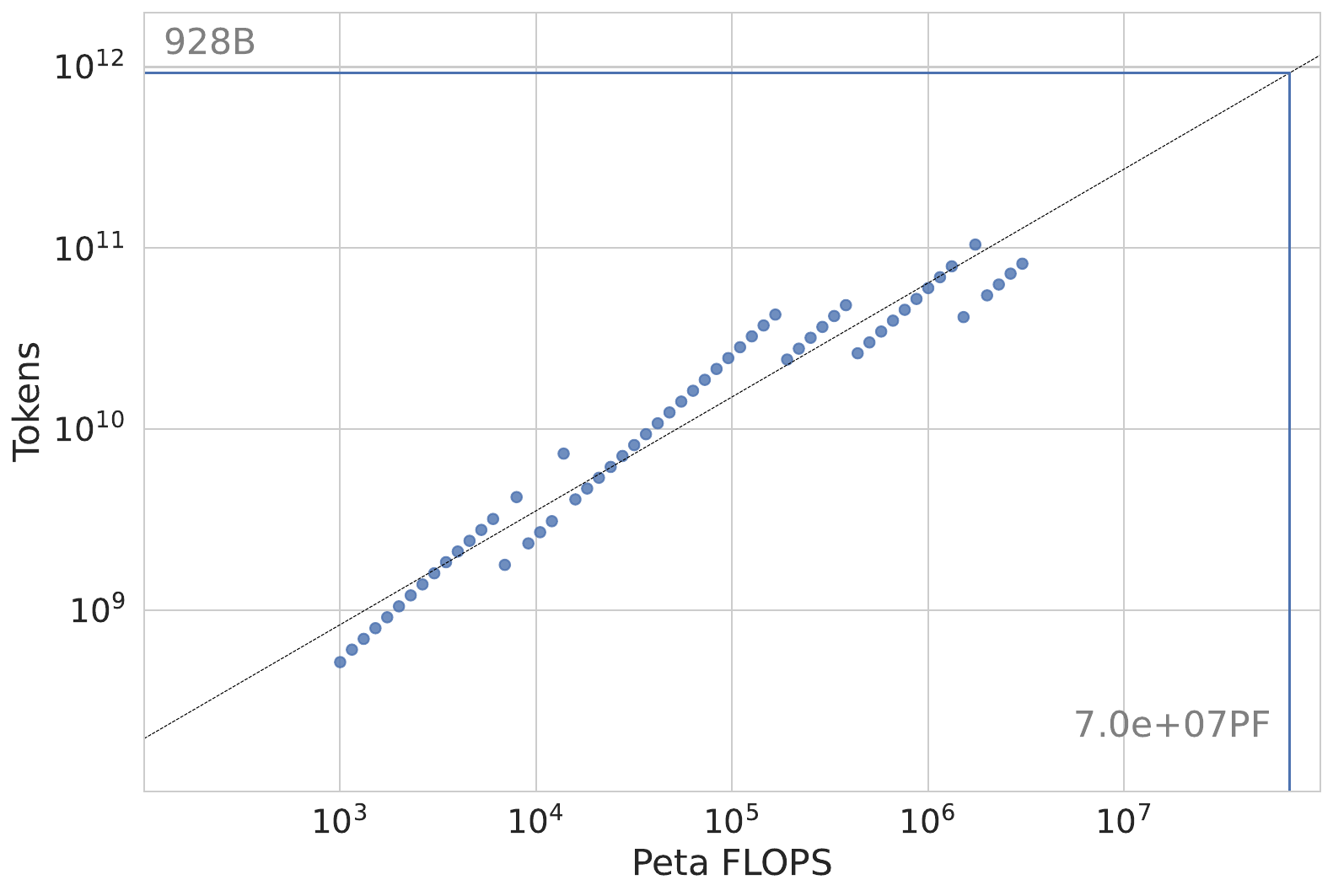}
        \caption{T2X(V) Power law of $C$ and $D$}
        \label{fig:video-computation-vs-token}
    \end{subfigure}
    \caption{Scaling laws of DiT-T2X model family. On the top-left (a) we show the loss curves of the T2X(I) model on a log-log scale for a range of model sizes from 92M to 6.6B. We follow~\cite{hoffmann2022training} to plot the envelope in gray points, which are used to estimate the power-law coefficients of the amount of computation ($C$) vs model parameters ($N$) (b) and the computation vs tokens ($D$) (c). Based on the scaling law of the T2X(I) model, we plot the scaling law of the corresponding T2X(V) model in (d), (e), and (f).}
    \label{fig:image_scaling_laws}
\end{figure}

In contrast to prior scaling laws on large language models~\cite{kaplan2020scaling,hoffmann2022training,touvron2023llama,achiam2023gpt,anil2023palm} and image generation models~\cite{li2024scalability,kilian2024computational}, video generation models typically rely on pre-trained image models. Consequently, our initial step involved establishing the foundational scaling laws pertinent to text-to-image. Building upon these foundational scaling laws, we subsequently derived the scaling laws applicable to the text-to-video model. By integrating these two sets of scaling laws, we were able to systematically determine the appropriate model and data configuration for video generation tasks. 


\subsubsection{Image model scaling law}

Kaplan et.al~\cite{kaplan2020scaling} and Hoffmann et.al~\cite{hoffmann2022training} explored emperical scaling laws for language models on cross-entropy loss. In the field of diffusion based visual generation, Li et.al~\cite{li2024scalability} study the scaling properties on UNet, while transformer based works such as DiT~\cite{peebles2023scalable}, U-ViT~\cite{bao2023all}, Lumina-T2X~\cite{gao2024lumina}, and SD3~\cite{esser2024scaling} only study the scaling behavior between sample quality and network complexity, leaving the power-laws about the computation resources and MSE loss used by diffusion models unexplored.

In order to fill the gap, we develop a family of DiT-like models, named as DiT-T2X to distinguish from the original DiT, where X can be the image (I) or the video (V).
DiT-T2X applies T5-XXL~\cite{raffel2020exploring} as the text encoder and the aformentioned 3D VAE as the image encoder. The text information is injected to the model according to cross-attention layers. The DiT-T2X family has seven sizes ranging from 92M to 6.6B. The models were trained using DDPM~\cite{ho2020denoising} and v-prediction~\cite{salimans2022progressive} with consistent hyperparameters and the same dataset with 256px resolution. We follow the experiment method introduced by~\cite{hoffmann2022training} and build the neural scaling laws to fit 
\begin{equation}\label{eq:scaling_law}
N_{opt}=a_1C^{b_1},\quad D_{opt}=a_2C^{b_2}.    
\end{equation}

As shown in Fig.~\ref{fig:image_scaling_laws} (a), the loss curve of each model decreases from top left to bottom right, and it always passes through the loss curve of the larger size model adjacent to it. It means that each curve will form two intersections with curves of the larger and the smaller models. Under the corresponding computation resources between the two intersections, the middle-sized model is optimal (with the lowest loss). After obtaining the envelope of lowest losses across all the x-axis values, we fill the Equation ~\eqref{eq:scaling_law} to find out that $a_1=5.48\times10^{-4}, b_1=0.5634, a_2=0.324$ and $b_2=0.4325$, where the units of $a_1, a_2, N_{opt}, D_{opt}$ are billions while $C$ has a unit of Peta FLOPs. The Fig.~\ref{fig:image_scaling_laws} (b) and Fig.~\ref{fig:image_scaling_laws} (c) show that DiT-T2X(I) family fits the power law quite well. Finally, given computation budgets, we can calculate the optimal model size and dataset size. 

\subsubsection{Video model scaling law}
Based on the scaling law of the T2X(I) model, we select the optimal image checkpoint (\textit{i.e.,} the model on the envelope) corresponding to each size model to serve as the initialization model for the video scaling law experiment. Fig.~\ref{fig:image_scaling_laws} (d), Fig.~\ref{fig:image_scaling_laws} (e), and Fig.~\ref{fig:image_scaling_laws} (f) illustrate the scaling law results of the T2X(V) model, where $a_1=0.0189, b_1=0.3618, a_2=0.0108$ and $b_2=0.6289$.
Based on the results of Fig.~\ref{fig:image_scaling_laws} (b) and Fig.~\ref{fig:image_scaling_laws} (e), and taking into account the training consumption and inference cost, we finally set the model size to 13B. Then the number of tokens for image and video training can be calculated as shown in Fig.~\ref{fig:image_scaling_laws} (c) and Fig.~\ref{fig:image_scaling_laws} (f). It is worth noting that the amount of training tokens calculated by image and video scaling laws is only related to the first stage of training for images and videos respectively. The scaling property of progressive training from low-resolution to high-resolution will be left explored in future work.


%% file: tax/pre-training.tex
\subsection{Model-pretraining}\label{model-pretrain}
We use Flow Matching~\cite{lipman2022flow} for model training and split the training process into multiple stages. We first pretrain our model on 256px and 512px images, then conduct joint training on images and videos from 256px to 960px. 

\subsubsection{Training Objective}
In this work, we employ the Flow Matching framework~\cite{lipman2022flow,esser2024scaling,chen2018neural} to train our image and video generation model. Flow Matching transforms a complex probability distribution into a simple probability distribution through a series of variable transformations of the probability density function, and generates new data samples through inverse transformations.

During the training process, given an image or video latent representation ${\rm \bf{x}}_1$ in the training set. We first sample $t\in [0,1]$  from a logit-normal distribution~\cite{esser2024scaling} and initialize a noise ${\rm \bf{x}}_0 \sim \mathcal{N}({\rm \bf{0}}, {\rm \bf{I}})$ following Gaussion distribution. The training sample ${\rm \bf{x}}_t$ is then constructed using a linear interpolation method~\cite{lipman2022flow}. The model is trained to predict the velocity ${\rm \bf{u}}_t=d {\rm \bf{x}}_t/dt$, which guides the sample ${\rm \bf{x}}_t$ towards the sample ${\rm \bf{x}}_1$. The model parameters are optimized by minimizing the mean squared error between the predicted velocity ${\rm \bf{v}}_t$ and the ground truth velocity ${\rm \bf{u}}_t$, expressed as the loss function
\begin{equation}
  \label{eq:fm}
  \mathcal{L}_{{\rm generation}}=\mathbb{E}_{t,{\rm \bf{x}}_0,{\rm \bf{x}}_1}\|{\rm \bf{v}}_t - {\rm \bf{u}}_t  \|^2.
\end{equation}

During the inference process, a noise sample ${\rm \bf{x}}_0 \sim \mathcal{N}({\rm \bf{0}}, {\rm \bf{I}})$ is drawn initially. The first-order Euler ordinary differential equation (ODE) solver is then used to compute ${\rm \bf{x}}_1$ by integrating the model's estimated values for $d{\rm \bf{x}}_t/dt$. This process ultimately generates the final sample ${\rm \bf{x}}_1$.

\subsubsection{Image Pre-training}

At our early experiments, we found that a well pretrained model significantly accelerates the convergence of video training and improves video generation performance. Therefore, we introduce a two-stage progressive image pretraining strategy to serve as a warmup for video training. 

\textbf{Image stage 1 (256px training).} The model is first pretrained with low-resolution 256px images. Specifically, we follow previous work~\cite{podell2023sdxl} to enable multi-aspect training based on 256px, which helps the model learn to generate images with a wide range of aspect ratios while avoiding the text-image misalignments caused by the crop operation in image preprocessing. Meanwhile, pretraining with low resolution samples allows the model to learn more low-frequency concepts from a larger amount of samples. 

\textbf{Image stage 2 (mix-scale training).} We introduce a second image-pretraining stage to further facilitate the model ability on higher resolutions, such as 512px. A trivial solution is to directly funetuning on images based on 512px. However, we found that the model performance finetuned on 512px images will degrade severely on 256px image generation, which may affect the following video pretraining on 256px videos. Therefore, we propose \textit{mix-scale training}, where two or more scales of multi-aspect buckets are included for each training global batch. Each scale have an anchor size, and then the multi-aspect buckets are built based on the anchor size. We train the model on a two-scale dataset with anchor sizes 256px and 512px for learning higher resolution images while maintaining the ability on low resolutions. We also introduce dynamic batch sizes for micro batches with different image scales, maximaizing the GPU memory and computation utilization.

\subsubsection{Video-Image joint training}

\textbf{Multiple aspect ratios and durations bucketization.}  After the data filtering process described in Section \ref{data_filtering}, the videos have different aspect ratios and durations. To effectively utilize the data, we categorize the training data into buckets based on duration and aspect ratio. We create $B_T$ duration buckets and $B_{AR}$ aspect ratio buckets, resulting in a total of $B_T \times B_{AR}$ buckets. As the number of tokens varies across buckets, we assign each bucket a maximum batch size that can prevent out-of-memory (OOM) errors, to optimize GPU resource utilization. Before training, all data is allocated to the nearest bucket. During training, each rank randomly pre-fetches batch data from a bucket. This random selection ensures the model is trained on varying data sizes at each step, which helps maintain model generalization by avoiding the limitations of training on a single size.

\textbf{Progressive Video-Image Joint Training.} Generating high-quality, long-duration video sequences directly from text often leads to difficulties in model convergence and suboptimal results. Therefore, progressive curriculum learning has become a widely adopted strategy for training text-to-video models. In \nameofmethod{}, we designed a comprehensive curriculum learning strategy, starting with model initialization using T2I parameters and progressively increasing video duration and resolution.

\begin{itemize}
\item \textbf{Low-resolution, short video stage.} The model establishes the basic mapping between text and visual content, ensuring consistency and coherence in short-term actions.

\item \textbf{Low-resolution, long video stage.} The model learns more complex temporal dynamics and scene changes, ensuring temporal and spatial consistency over a longer duration.

\item \textbf{High-resolution, long video stage.} The model enhances video resolution and detail quality while maintaining temporal coherence and managing complex temporal dynamics.
\end{itemize}

Additionally, at each stage, we incorporate images in varying proportions for video-image joint training. This approach addresses the scarcity of high-quality video data, enabling the model to learn more extensive and diverse world knowledge. It also effectively prevents catastrophic forgetting of image-space semantics due to distributional discrepancies between video and image data.

\subsection{Prompt Rewrite} 
To address the variability in linguistic style and length of user-provided prompts, we employ the Hunyuan-Large model \cite{sun2024hunyuanlargeopensourcemoemodel} as our prompt rewrite model to adapt the original user prompt to the model-preferred prompt. Utilized within a training-free framework, the prompt rewrite model capitalizes on detailed prompt instructions and in-context learning examples to enhance its performance. The key functionalities of this prompt rewrite module are as follows:

\begin{itemize}
    \item \textbf{Multilingual Input Adaptation}: The module is designed to process and comprehend user prompts across various languages, ensuring that meaning and context are preserved.
    \item \textbf{Standardization of Prompt Structure}: The module rephrases prompts to conform to a standardized information architecture, akin to training captions.
    \item \textbf{Simplification of Complex Terminology}: The module simplifies complex user wording into more straightforward expressions, all while maintaining the user's original intent.
\end{itemize}

Furthermore, we implement a self-revision technique \cite{kim2024reex} to refine the final prompt. This involves a comparative analysis between the original prompt and the rewritten version, ensuring that the output is both accurate and aligned with the model's capabilities.

To accelerate and simplify the application process, we also fine-tune a Hunyuan-Large model with LoRA for prompt rewriting. The training data for this LoRA tuning was sourced from the high-quality rewrite pairs collected through the training-free method.

\subsection{High-performance Model Fine-tuning}\label{sft}
In the pre-training stage, we utilized a large dataset for model training. While this dataset is rich in information, it displayed considerable variability in data quality. To create a robust generation model capable of producing high-quality, dynamic videos and improving its proficiency in continuous motion control and character animation, we carefully selected four specific subsets from the full dataset for fine-tuning.
These subsets underwent an initial screening using automated data filtering techniques, followed by manual review. Additionally, we implemented various model optimization strategies to maximize generation performance.



%% file: tax/accelerate.tex
\section{Model Acceleration}
\label{sec:accelerate}

\begin{figure}[htbp]
    \centering
    \begin{subfigure}[b]{0.28\textwidth} 
        \centering
        \includegraphics[width=\textwidth]{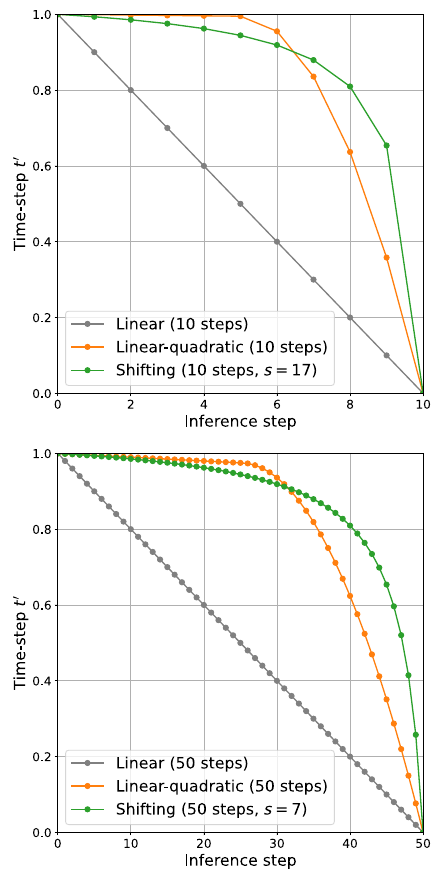}
        \caption{}
        \label{fig:shifting}
    \end{subfigure}
    \hfill
    \begin{subfigure}[b]{0.69\textwidth} 
        \centering
        \includegraphics[width=\textwidth]{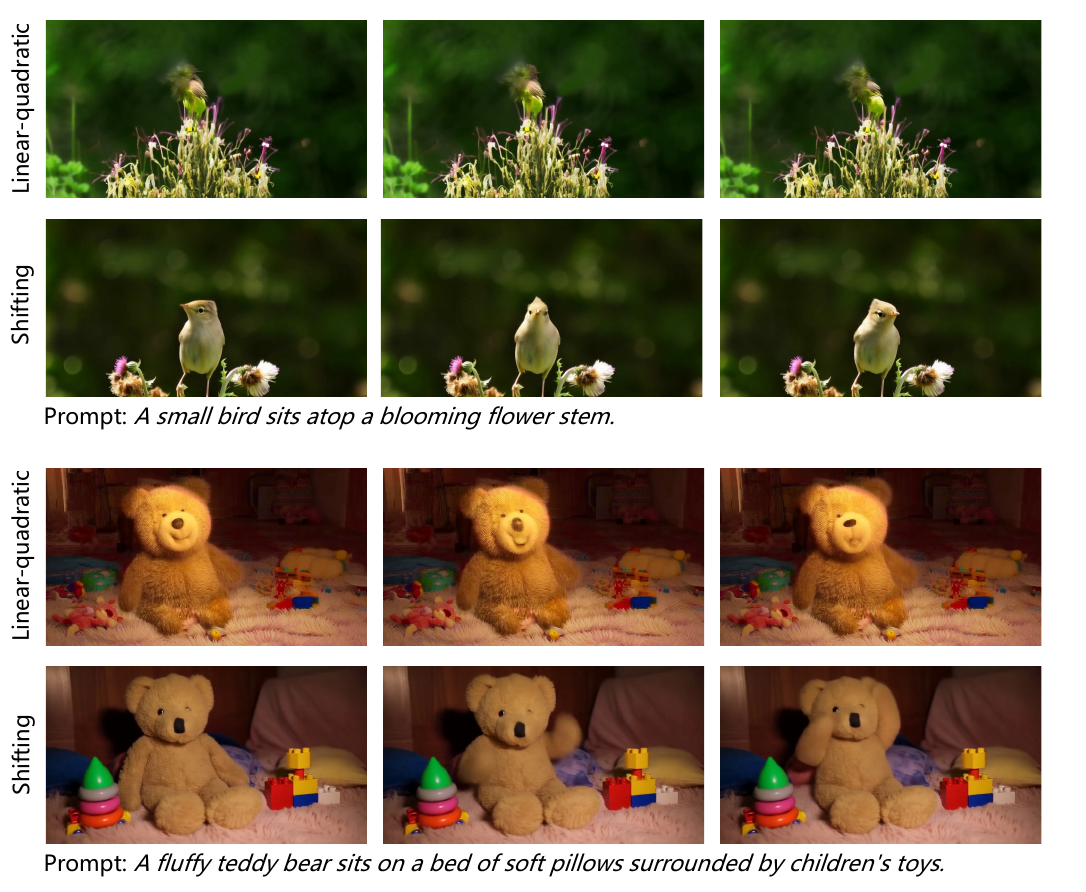}
        \caption{}
        \label{fig:shifting_example}
    \end{subfigure}
    \caption{(a) Different time-step schedulers. For our shifting stragty, we set a larger shifting factor $s$ for a lower inference step. (b) Generated videos with only 10 inference steps. The shifting stragty leads to significantly better visual quality.}
    \label{fig:overall}
\end{figure}

\subsection{Inference Step Reduction }

To improve the inference efficiency, we firstly consider reducing the number of inference steps. Compared to image generation, it is more challenging to maintain the spatial and temporal quality of the generated videos with lower inference steps. Inspired by a previous observation that the first time-steps contribute to most changes during the generation process \cite{zhao2024real,polyak2024movie,zhang2022unsupervised,zhang2023shiftddpms}, we utilize the time-step shifting to handle the case of lower inference steps. Specifically, given the inference step $q \in \{1,2,...,Q\}$, $t=1-\frac{q}{Q}$ is the input time condition for the generation model, where the noise is initialized at $t=1$ and the generation process halts at $t=0$. Instead of using $t$ directly, we map $t$ to $t'$ with a shifting function $t'=\frac{s*t}{1+(s-1)*t}$, where $t'$ is the input time condition and $s$ is the shifting factor. If $s>1$, the flow model is conditioned more on early time steps. A critical observation is that a lower inference step requires a larger shifting factor $s$. Empirically, $s$ is set as 7 for 50 inference steps, while $s$ should be increased to 17 when the number of inference steps is smaller than 20. The time-step shifting strategy enables the generation model to match the results of numerous inference steps with a reduced number of steps.

MovieGen \cite{polyak2024movie} applies the linear-quadratic scheduler to achieve a similar goal. The schedulers are visualized in Figure \ref{fig:shifting}. However, we find that our time-step shifting is more effective than the linear-quadratic scheduler in the case of extremely low number of inference steps, $e.g.$, 10 steps. As shown in Figure \ref{fig:shifting_example}, the linear-quadratic scheduler results in worse visual quality.

\subsection{Text-guidance Distillation}

Classifier-free guidance (CFG) \cite{ho2022classifier} significantly improves the sample quality and motion stability of text-conditioned diffusion models.
However, it increases computational cost and inference latency since it requires extra output for the unconditional input at each inference step.
Especially for the large video model and high-resolution video generation, the inference burden is extremely expensive when generating text-conditional and text-unconditional videos, simultaneously.
To solve this limitation, we distill the combined output for unconditional and conditional inputs into a single student model \cite{meng2023distillation}.
Specifically, the student model is conditioned on a guidance scale and shares the same structures and hyper-parameters as the teacher model.
We initialize the student model with the same parameters as the teacher model and train with the guidance scale randomly sampled from 1 to 8. We experimentally find that text-guidance distillation approximatively brings 1.9x acceleration. 

\subsection{Efficient and Scalable Training}

To achieve scalability and efficient training, 
we train \nameofmethod{} on AngelPTM \cite{nie2023angel}, 
the large-scale pretraining framework from Tencent Angel machine learning team.
In this part, we first outline the hardware and infrastructure used for training, and then give a detailed introduction to the model parallel method and its optimization methods, followed by the automatic fault tolerance mechanism.

\subsubsection{Hardware Infrastucture}
To ensure efficient communication in large-scale distributed training, we setup a dedicated distributed training framework termed Tencent XingMai network \cite{2024tccl} for highly efficient inter-server communication. 
The GPU scheduling for all training tasks is completed through the Tencent Angel machine learning platform, which provides powerful resource management and scheduling capabilities.

\subsubsection{Parallel Strategy}
\nameofmethod{} training adopts 5D parallel strategies, including tensor parallelism (TP) \cite{2019megatron}, sequence parallelism (SP) \cite{2022sequence-parallelism}, context parallelism (CP) \cite{2024context-parallelism}, and data parallelism combined with Zero optimization (DP + ZeroCache \cite{nie2023angel}).
The tensor parallelism (TP) is based on the principle of block calculation of matrices. The model parameters (tensors) are divided into different GPUs to reduce GPU memory usage and accelerate the calculation. Each GPU is responsible for the calculation of different parts of tensors in the layer.

Sequence parallelism (SP) is based on TP. The input sequence dimension is sliced to reduce the repeated calculation of operators such as LayerNorm and Dropout, and reduce the storage of the same activations, which effectively reduces the waste of computing resources and GPU memory.
In addition, for input data that does not meet the SP requirements, the engineering equivalent SP Padding capability is supported.

Context parallelism (CP) is sliced in the sequence dimension to support long-sequence training. Each GPU is responsible for calculating the Attention of different sequence slices. Specifically, Ring Attention \cite{2023ring-attention} is used to achieve efficient training of long sequences through multiple GPUs, breaking through the GPU memory limitation of a single GPU.

In addition, data parallelism + ZeroCache is leveraged to support horizontal expansion through data parallelism to meet the demand for increasing training data sets. Then, based on data parallelism, the ZeroCache optimization strategy is used to further reduce the redundancy of model states (model parameters, gradients, and optimizer states), and we unify the use of GPU memory to maximize the GPU memory usage efficiency.

\subsubsection{Optimization}
\textbf{Attention optimization.} 
As the sequence length increases, the attention calculation becomes the main bottleneck of training. We accelerated the attention calculation with FusedAttention.

\textbf{Recomputation and activations offload optimization.} Recomputation is a technology that trade calculations for storage. 
It is mainly made up of three parts: a) specifying certain layers or blocks for recalculation, b) releasing the activations in the forward calculation, and c) obtaining the dependent activations through recalculation in the backward calculation, which significantly reduces the use of GPU memory during training.
In addition, considering the PCIe bandwidth and the host memory size, a layer-based activation offload strategy is adopted. Without reducing the training performance, the activations in the GPU memory are offloaded to the host memory, further saving GPU memory.

\subsubsection{Automatic fault tolerance}
In terms of the large-scale training stability of \nameofmethod{}, an automatic fault tolerance mechanism is used to quickly restore training for common hardware failures.
This avoids frequent occurrence of the manual recovery of training tasks.
By automatically detecting errors and quickly replacing healthy nodes to pull up training tasks, the training stability is 99.5\%.

%% file: exp/text_encoder.tex
\begin{figure}[h]
    \centering
    \ifhq
    \includegraphics[trim={0cm 5cm 0cm 4cm},clip,width=\linewidth]{hqfigures/text_alignment.pdf}
    \else
    \includegraphics[width=\linewidth]{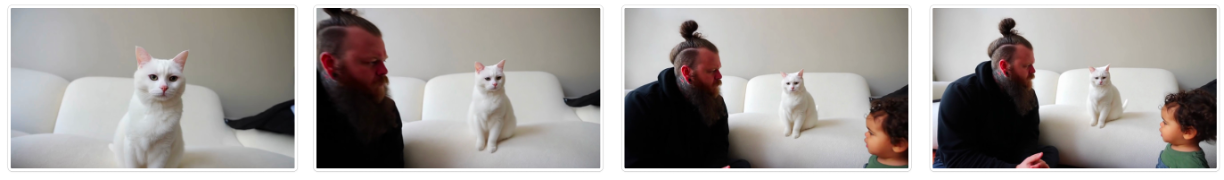}
    \fi
    \caption{{Prompt: A white cat sits on a white soft sofa like a person, while its long-haired male owner, with his hair tied up in a topknot, sits on the floor, gazing into the cat's eyes. His child stands nearby, observing the interaction between the cat and the man. } }
    \label{fig:text-align}
\end{figure}

\myPara{Text Alignment}
One of the key metrics for video generative models is their ability to follow text prompts accurately. This capability is essential for the effectiveness of these models. However, some open-source models often struggle to capture all subjects or accurately represent the relationships between multiple subjects, particularly when the input text prompt is complex.
\nameofmethod{} demonstrates robust capabilities in generating videos that closely adhere to the provided text prompts. As illustrated in Figure \ref{fig:text-align}, it effectively manages multiple subjects within the scene.

%% file: exp/sft.tex
\begin{figure}[!htbp]
    \centering
    \begin{subfigure}{\textwidth}
        \centering
        \ifhq
        \includegraphics[width=\textwidth]{hqfigures/high-quality-1.pdf}
        \else
        \includegraphics[width=\textwidth]{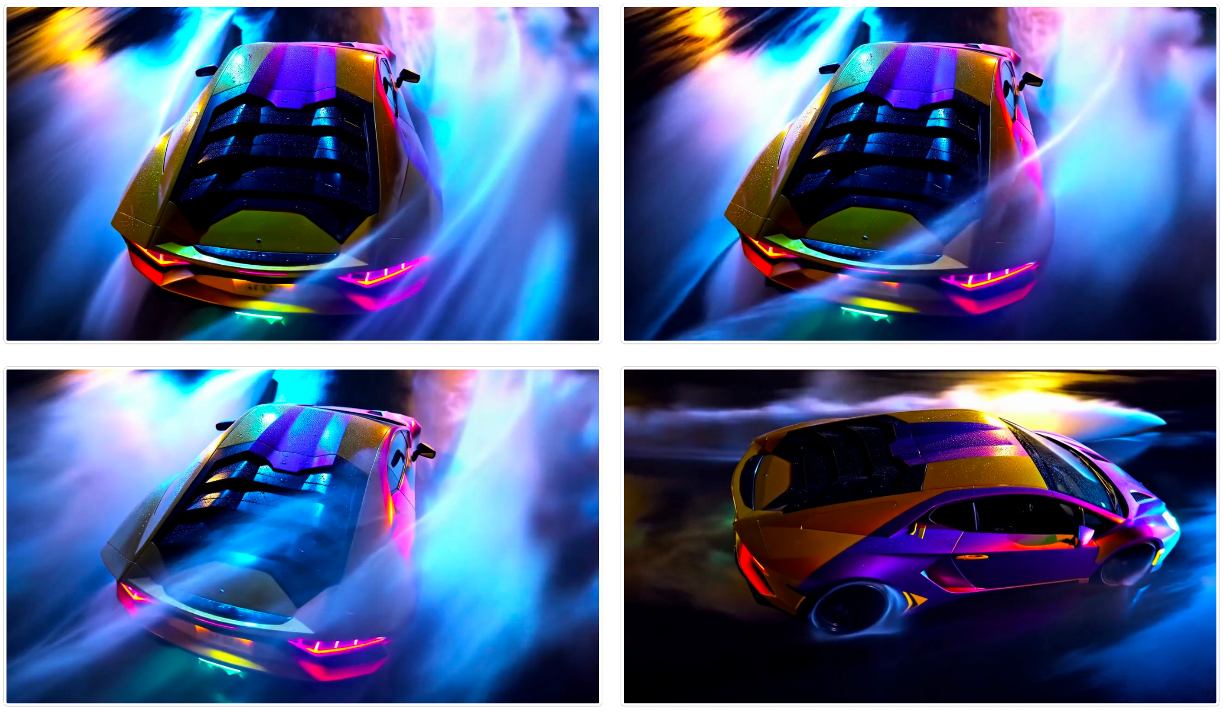}
        \fi
        \caption{Prompt: the ultra-wide-angle lens follows closely from the hood, with raindrops continuously splattering against the lens. Ahead, a sports car speeds around a corner, its tires violently skidding against the wet road, creating a mist of water. Neon lights refract in the rain, leaving colorful streaks on the car's surface. The camera swiftly shifts to the side of the car, capturing the wheels spinning at high speed, before finally moving to the rear.}
    \end{subfigure}
    \hfill
    \begin{subfigure}{\textwidth}
        \centering
        \ifhq
        \includegraphics[width=\textwidth]{hqfigures/high-quality-2.pdf}
        \else
        \includegraphics[width=\textwidth]{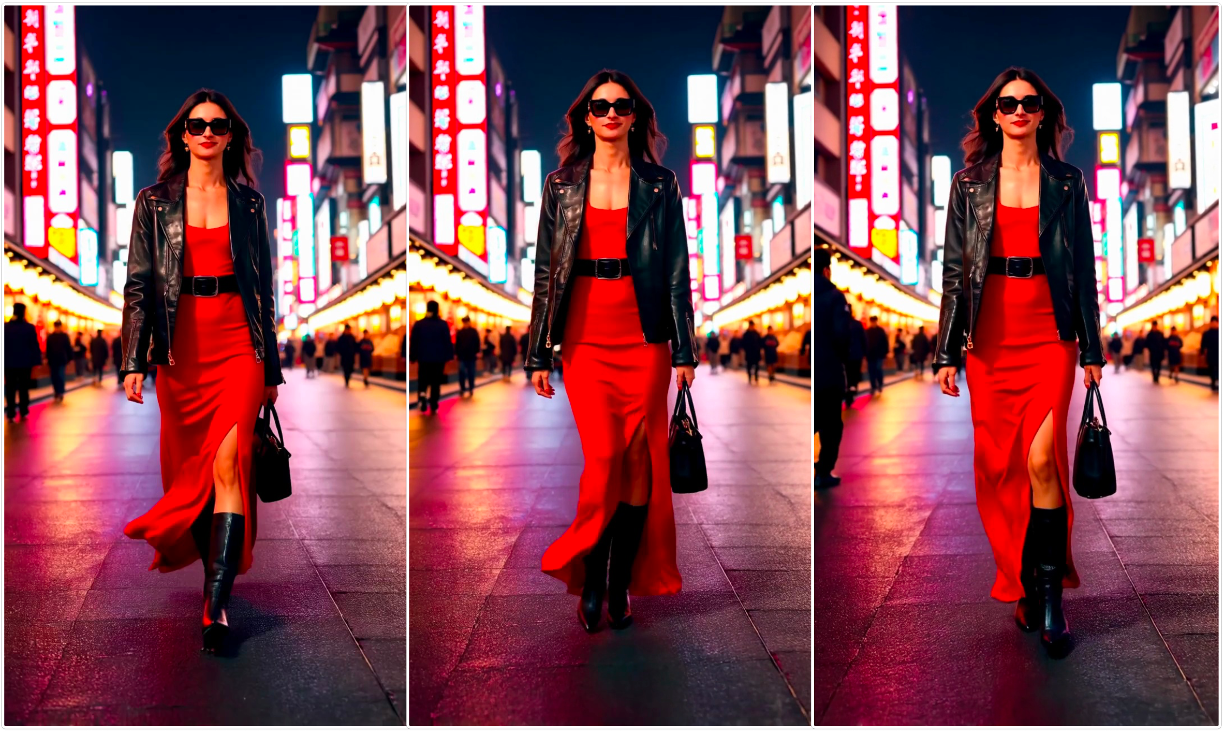}
        \fi
        \caption{Prompt: a stylish woman walks down a Tokyo street filled with warm glowing neon and animated city signage. She wears a black leather jacket, a long red dress, and black boots, and carries a black purse. She wears sunglasses and red lipstick. She walks confidently and casually. The street is damp and reflective, creating a mirror effect of the colorful lights. Many pedestrians walk about.}
    \end{subfigure}
    \caption{High-quality videos generated by HunyuanVideo.}
    \label{fig:high_quality}
\end{figure}

\myPara{High-quality}
We also perform a fine-tuning process to enhance the spatial quality of the generated videos. As illustrated in Figure \ref{fig:high_quality}, \nameofmethod{} is capable of producing videos with ultra-detailed content.

\begin{figure}[!htbp]
    \centering
    \begin{subfigure}{\textwidth}
        \centering
        \ifhq
        \includegraphics[trim={4.8cm 4.8cm 4.8cm 4.8cm},clip,width=0.95\textwidth]{hqfigures/high_motion_3.pdf}
        \else
        \includegraphics[width=\textwidth]{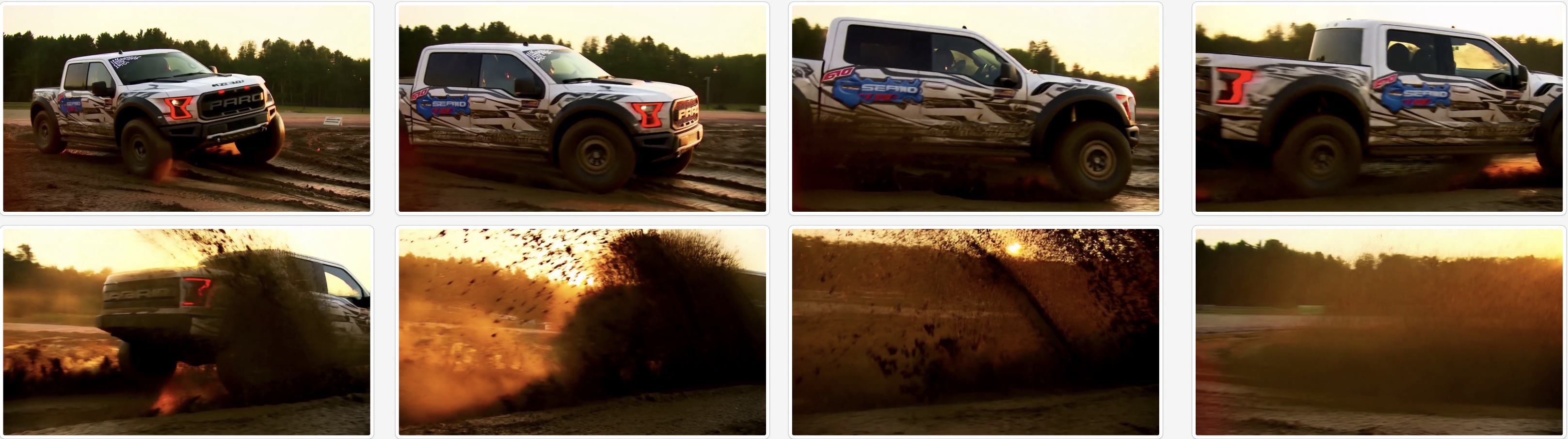}
        \fi
        \captionsetup{font=small}
        \caption{Prompt: At sunset, a modified Ford F-150 Raptor roared past on the off-road track. The raised suspension allowed the huge explosion-proof tires to flip freely on the mud, and the mud splashed on the roll cage.}
        \label{fig:hm_1}
    \end{subfigure}
    \hfill
    \begin{subfigure}{\textwidth}
        \centering
        \ifhq
        \includegraphics[trim={4.8cm 4.8cm 4.8cm 4.8cm},clip,width=0.95\textwidth]{hqfigures/high_motion_5.pdf}
        \else
        \includegraphics[width=\textwidth]{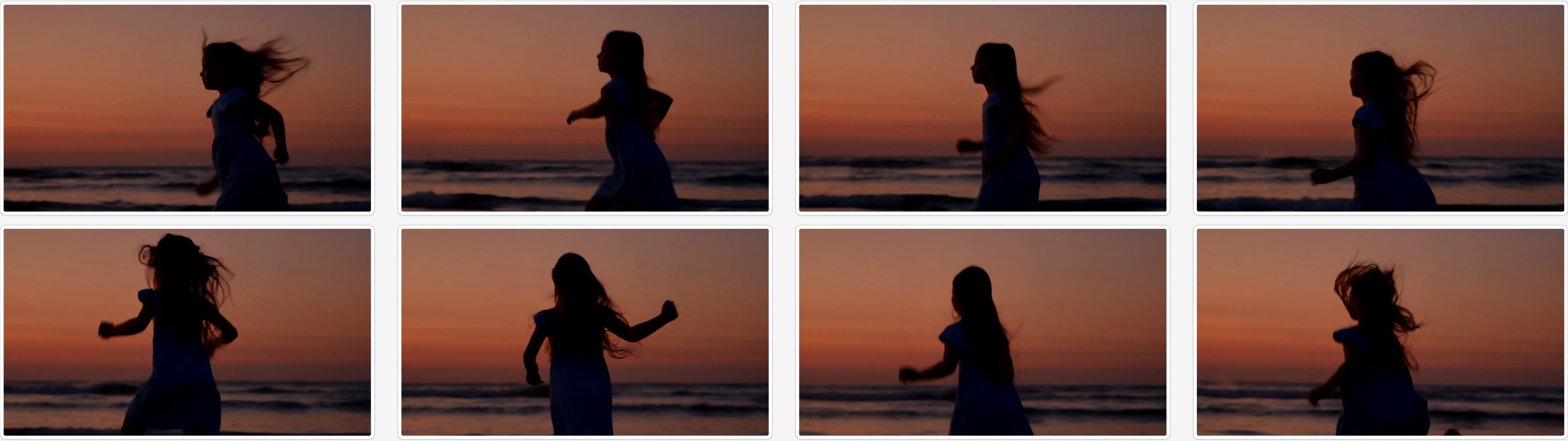}
        \fi
        \captionsetup{font=small}
        \caption{Prompt: The panning camera moves forward slowly, with a depth of field in the middle focus, and warm sunset light covers the screen. The girl in the picture runs with her skirt fluttering, turns and jumps.}
        \label{fig:hm_2}
    \end{subfigure}
    \hfill
    \begin{subfigure}{\textwidth}
        \centering
        \ifhq
        \includegraphics[trim={4.8cm 4.8cm 4.8cm 4.8cm},clip,width=0.95\textwidth]{hqfigures/high_motion_2.pdf}
        \else
        \includegraphics[width=\textwidth]{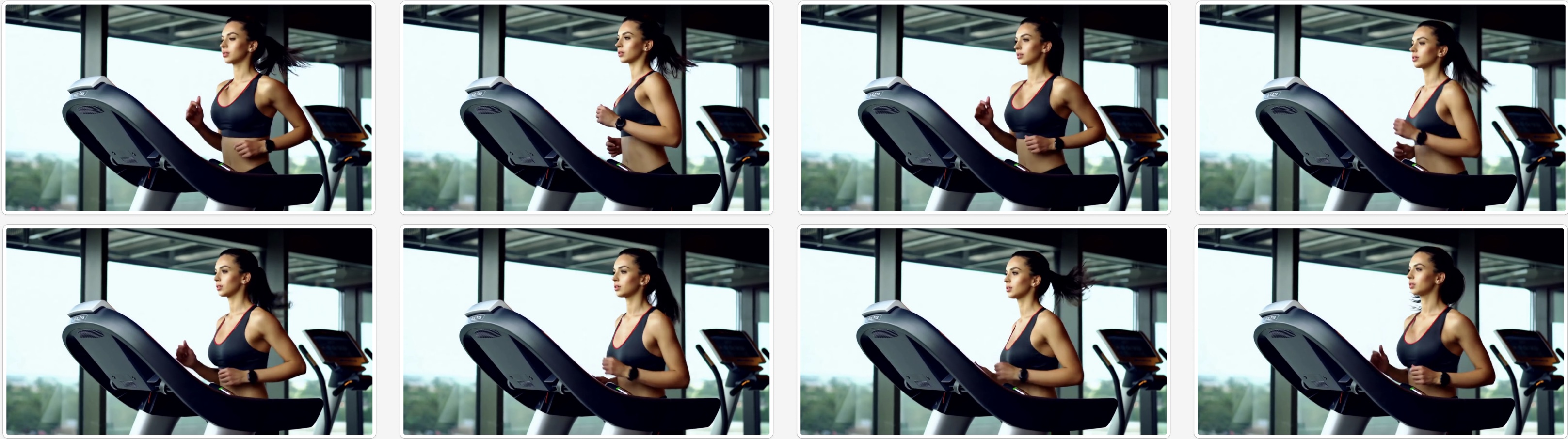}
        \fi
        \captionsetup{font=small}
        \caption{Prompt: In the gym, a woman in workout clothes runs on a treadmill. Side angle. Realistic, Indoor lighting, Professional.}
        \label{fig:hm_3}
    \end{subfigure}
    \hfill
    \begin{subfigure}{\textwidth}
        \centering
        \ifhq
        \includegraphics[trim={4.8cm 4.8cm 4.8cm 4.8cm},clip,width=0.95\textwidth]{hqfigures/high_motion_7.pdf}
        \else
        \includegraphics[width=\textwidth]{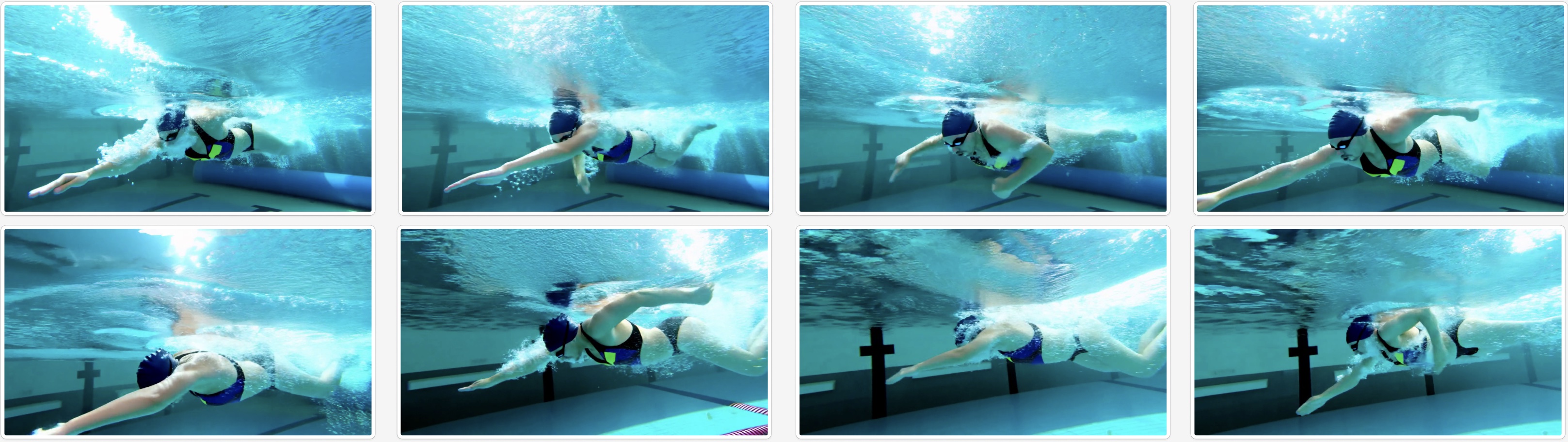}
        \fi
        \captionsetup{font=small}
        \caption{Prompt: Swimmer swimming underwater, in slow motion. Realistic, Underwater lighting, Peaceful.}
        \label{fig:hm_4}
    \end{subfigure}
    \hfill
    \begin{subfigure}{\textwidth}
        \centering
        \ifhq
        \includegraphics[trim={4.8cm 4.8cm 4.8cm 4.8cm},clip,width=0.95\textwidth]{hqfigures/high_motion_8.pdf}
        \else
        \includegraphics[width=\textwidth]{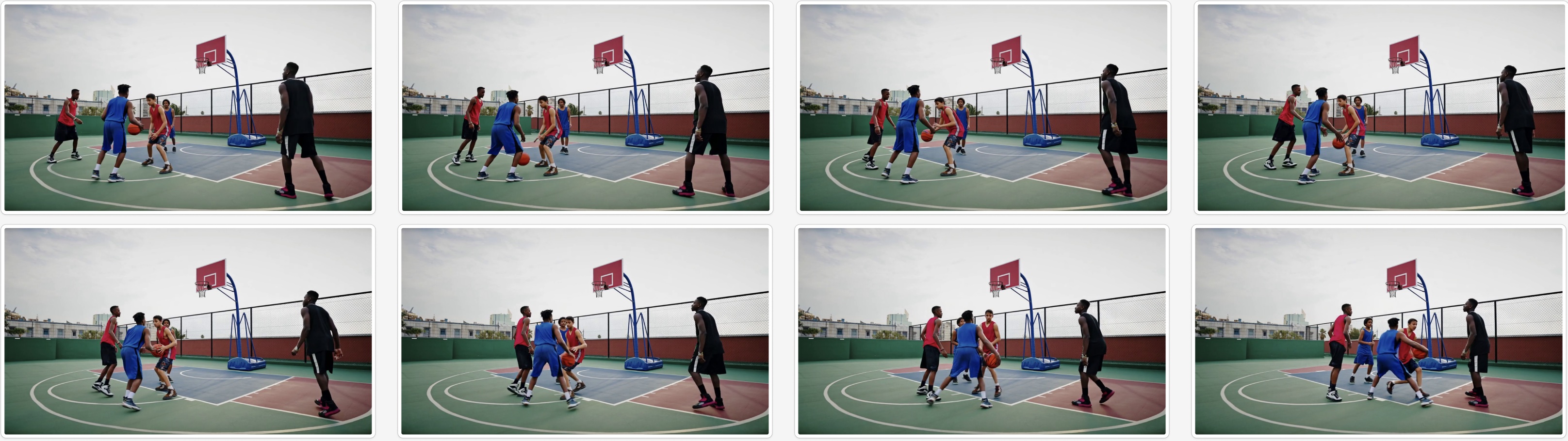}
        \fi
        \captionsetup{font=small}
        \caption{Prompt: On the rooftop, there is an open-air basketball court, and five male students are playing basketball. Realistic, Natural lighting, Casual.}
        \label{fig:hm_5}
    \end{subfigure}
    \caption{High-motion dynamics videos generated by \nameofmethod{}.}
    \label{fig:high_motion}
\end{figure}

\myPara{High-motion Dynamics}
In this part, we demonstrate \nameofmethod{}'s capabilities in producing high-dynamic videos based on given prompts. As shown in Figure \ref{fig:high_motion}, our model excels in generating videos that encompass a wide range of scenes and various types of motion.

\begin{figure}[!htbp]
    \centering
    \ifhq
    \includegraphics[trim=0cm 5cm 0cm 4cm,clip,width=\linewidth]{hqfigures/concept.pdf}
    \else
    \includegraphics[width=\linewidth]{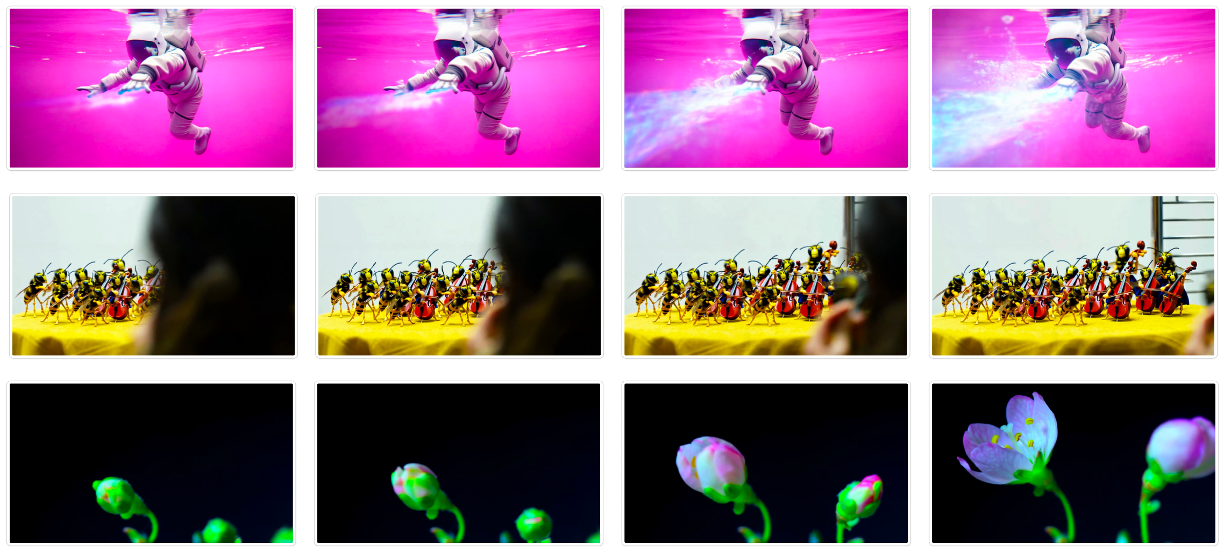}
    \fi
    \caption{\small {\nameofmethod{}'s performance on concept generalization. The results of the three rows correspond to the text prompts (1) `In a distant galaxy, an astronaut floats on a shimmering, pink, gemstone-like lake that reflects the vibrant colors of the surrounding sky, creating a stunning scene. The astronaut gently drifts on the lake's surface, the soft sounds of water whispering the planet's secrets. He reaches out, his fingertips gliding over the cool, smooth water. ', (2) `A macro lens captures a tiny orchestra of insects playing instruments.' and (3) `The night-blooming cactus flowers in the evening, with a brief, rapid closure. Time-lapse shot, extreme close-up. Realistic, Night lighting, Mysterious.' respectively.} }
    \label{fig:concept}
\end{figure}

\myPara{Concept Generalization}
One of the most desirable features of a generative model is its ability to generalize concepts. As illustrated in Figure \ref{fig:concept}, the text prompt describes a scene: "In a distant galaxy, an astronaut floats on a shimmering, pink, gemstone-like lake that reflects the vibrant colors of the surrounding sky, creating a stunning scene. The astronaut gently drifts on the lake's surface, while the soft sounds of water whisper the planet's secrets. He reaches out, his fingertips gliding over the cool, smooth water." Notably, this specific scenario has not been encountered in the training dataset. Furthermore, it is evident that the depicted scene combines several concepts that are also absent from the training data.

\begin{figure}[ht]
    \centering
    \ifhq
    \includegraphics[trim=0cm 5cm 0cm 4cm,clip,width=\linewidth]{hqfigures/sequential_motion.pdf}
    \else
    \includegraphics[width=\linewidth]{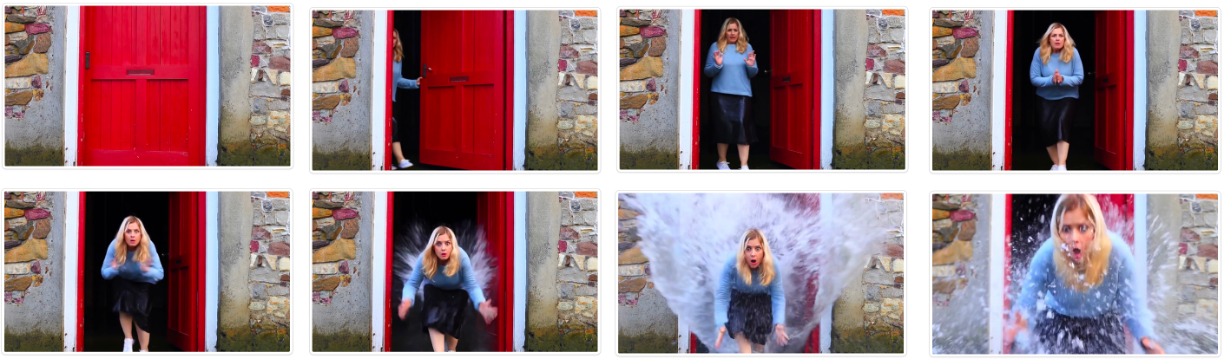}
    \fi
    \caption{{Prompt: The woman walks over and opens the red wooden door. As the door swings open, seawater bursts forth, in a realistic style.}}
    \label{fig:sequential-move}
\end{figure}

\myPara{Action Reasoning and Planning}
Leveraging the capabilities of large language models, \nameofmethod{} can generate sequential movements based on a provided text prompt. As demonstrated in Figure \ref{fig:sequential-move}, \nameofmethod{} effectively captures all actions in a photorealistic style.

\begin{figure}[!htbp]
    \centering
    \ifhq
    \includegraphics[trim=2cm 5cm 2cm 4cm,clip,width=\linewidth]{hqfigures/ocr.pdf}
    \else
    \includegraphics[width=\linewidth]{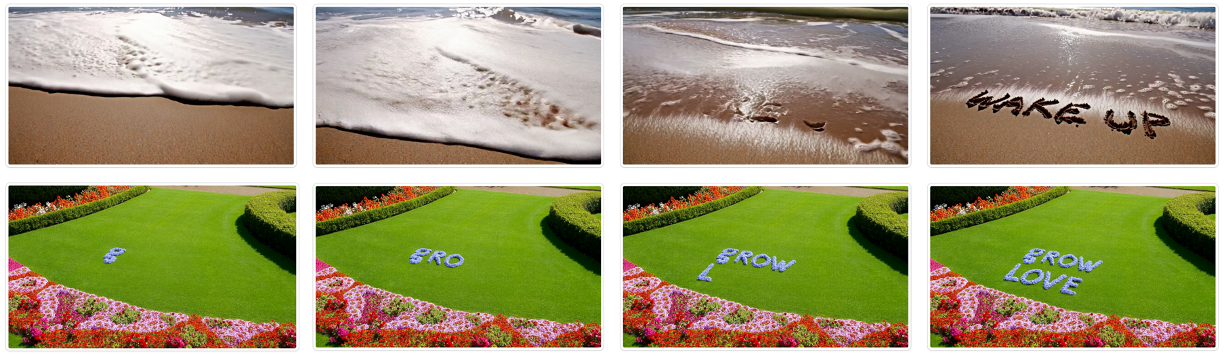}
    \fi
    \caption{High text-video alignment videos generated by HunyuanVideo. Top row: Prompt: A close-up of a wave crashing against the beach, the sea foam spells out ``WAKE UP'' on the sand. Bottom row: Prompt: In a garden filled with blooming flowers, ``GROW LOVE'' has been spelled out with colorful petals.}
    \label{fig:ocr}
\end{figure}

\myPara{Character Understanding and Writing}
\nameofmethod{} is capable of generating both scene text and gradually appearing handwritten text as shown in Fig.~\ref{fig:ocr}.

%% file: exp/human_eval.tex
\subsection{Comparison with SOTA Models}
To evaluate the performance of \nameofmethod{}, we selected five strong baselines from closed-source video generation models. In total, we utilized 1,533 text prompts, generating an equal number of video samples with \nameofmethod{} in a single run. For a fair comparison, we conducted inference only once, avoiding any cherry-picking of results. When comparing with the baseline methods, we maintained the default settings for all selected models, ensuring consistent video resolution. 60 professional evaluators performed the evaluation and the results are presented in Table \ref{tab:compare}. Videos were assessed based on three criteria: Text Alignment, Motion Quality, and Visual Quality. Notably, \nameofmethod{} demonstrated the best overall performance, particularly excelling in motion quality. We randomly sample 600 videos out of 1533 for public access\footnote{https://github.com/Tencent/HunyuanVideo}.
\begin{table}[h]
    \centering
    \footnotesize
    \caption{Model Performance Evaluation}
    \begin{tabular}{@{}llccccc@{}}
        \toprule
        Model Name                             &  Duration & Text Alignment  & Motion Quality  & Visual Quality  & Overall  & Ranking \\ \midrule
        \nameofmethod{} (Ours)               & 5s       & 61.8\%               & 66.5\%          & 95.7\%          & 41.3\%      & 1              \\
        CNTopA (API)       & 5s       & 62.6\%               & 61.7\%          & 95.6\%          & 37.7\%      & 2              \\
        CNTopB (Web)          & 5s       & 60.1\%               & 62.9\%          & 97.7\%          & 37.5\%      & 3              \\
        GEN-3 alpha (Web)           & 6s       & 47.7\%               & 54.7\%          & 97.5\%          & 27.4\%      & 4              \\
        Luma1.6 (API)                & 5s       & 57.6\%               & 44.2\%          & 94.1\%          & 24.8\%      & 5              \\
        CNTopC (Web)        & 5s       & 48.4\%               & 47.2\%          & 96.3\%          & 24.6\%      & 6              \\
         \bottomrule
    \end{tabular}
    \label{tab:compare}

\end{table}

%% file: tax/tv2a.tex
\subsection{Audio Generation based on Video}

Our video-to-audio(V2A) module is designed to enhance generated video content by incorporating synchronized sound effects and contextually appropriate background music. Within the conventional film production pipeline, Foley sound design constitutes an integral component, significantly contributing to the auditory realism and emotional depth of visual media. However, the creation of Foley audio is both time-intensive and demands a high degree of professional expertise. With the advent of an increasing number of text-to-video (T2V) models, most of them lack the corresponding foley generation capabilities, thereby limiting their ability to produce fully immersive content. Our V2A module addresses this critical gap by autonomously generating cinematic-grade foley audio tailored to the input video and textual prompts, thus enabling the synthesis of a cohesive and holistically engaging multimedia experience.

\subsubsection{Data }
Unlike text-to-video (T2V) models, video-to-audio (V2A) models have different requirements for data. As mentioned above, we constructed a video dataset comprising of video-text pairs. However, not all data in this dataset are suitable for training the V2A model. For example, some videos lack an audio stream, others contain extensive voice-over content or their ambient audio tracks have been removed and replaced with unrelated elements. To address these challenges and ensure data quality, we designed a robust data filtering pipeline specifically tailored for V2A training.

First, we filter out videos without audio streams or those in which the silence ratio exceeds 80\%. Next, we employ a frame-level audio detection model, like \cite{Hung2022}, to detect speech, music, and general sound in the audio stream. Based on this analysis, we classify the data into four distinct categories: \textit{pure sound}, \textit{sound with speech}, \textit{sound with music}, and \textit{pure music}. Subsequently, to prioritize high-quality data, we train a model inspired by CAVP \cite{luo2024diff} to compute a visual-audio consistency score, which quantifies the alignment between the visual and auditory components of each video. Using this scoring system in conjunction with the audio category labels, we systematically sample portions of data from each category, retaining approximately 250,000 hours from the original dataset for pre-training. For the supervised fine-tuning stage, we further refine our selection, curating a subset of millions of high-quality clips~(80,000 hours).

For feature extraction, we use CLIP \cite{clip} to obtain visual features at a temporal resolution of 4 fps and subsequently resample these features to align with the audio frame rate. To generate captions, we employ \cite{haji2024taming} as the sound captioning model and \cite{doh2023lp} as the music captioning model. When both sound and music captions are available, we merge them into a structured caption format, following the approach detailed in \cite{polyak2024movie}.

\begin{figure}[t]
    \centering
    \ifhq
    \includegraphics[trim={5cm 5cm 5cm 5cm},clip,width=0.9\linewidth]{hqfigures/vt2a_arch.pdf}
    \else
    \includegraphics[width=0.9\linewidth]{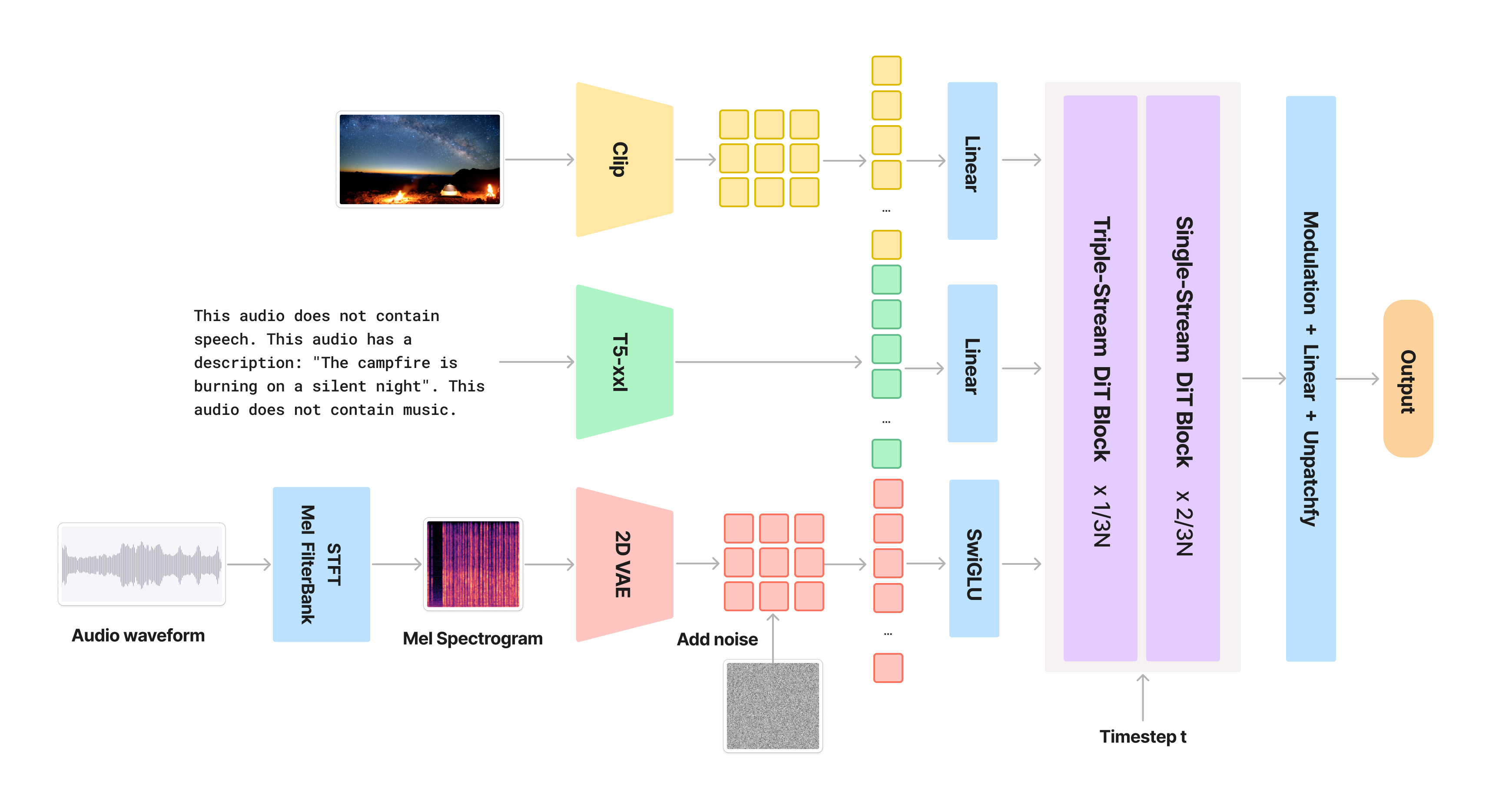}
    \fi
    \caption{The architecture of sound effect and music generation model. }
    \label{fig:audio-gen}
\end{figure}

\subsubsection{Model}
Just like the above-mentioned text-to-video model, our video-to-audio generation model also adopts a flow-matching-based diffusion transformer (DiT) as its architectural backbone. The detailed design of the model is depicted in Figure \ref{fig:audio-gen}, illustrating a transition from a triple-stream structure to a single-stream DiT framework.

The model operates within a latent space encoded by a variational autoencoder (VAE) trained on mel-spectrograms. Specifically, the audio waveform is first converted into a 2D mel-spectrogram representation. This spectrogram is subsequently encoded into a latent space using a pretrained VAE. For feature extraction, we leverage pretrained CLIP \cite{clip} and T5 \cite{raffel2020exploring} encoders to independently extract visual and textual features, respectively. These features are subsequently projected into the DiT-compatible latent space using independent linear projections followed by SwiGLU activation, as depicted in Figure \ref{fig:audio-gen}.

To effectively integrate multimodal information, we incorporate stacked triple-stream transformer blocks, which independently process visual, audio, and textual modalities. These are later followed by single-stream transformer blocks to ensure seamless fusion and alignment across modalities. This design enhances the alignment between audio-video and audio-text representations, facilitating improved multimodal coherence.

Once the latent representation is generated by the diffusion transformer, the VAE decoder reconstructs the corresponding mel-spectrogram. Finally, the mel-spectrogram is converted back into an audio waveform using a pre-trained HifiGAN vocoder \cite{kong2020hifi}. This framework ensures a high-fidelity reconstruction of audio signals while maintaining strong multimodal alignment.

%% file: applications/i2v.tex
\subsection{Hunyuan Image-to-Video}
\subsubsection{Pre-training}
\begin{figure}[htbp]
    \centering
    \ifhq
    \includegraphics[width=0.7\linewidth]{hqfigures/I2V_overall_token_replace.png}
    \else
    \includegraphics[width=0.7\linewidth]{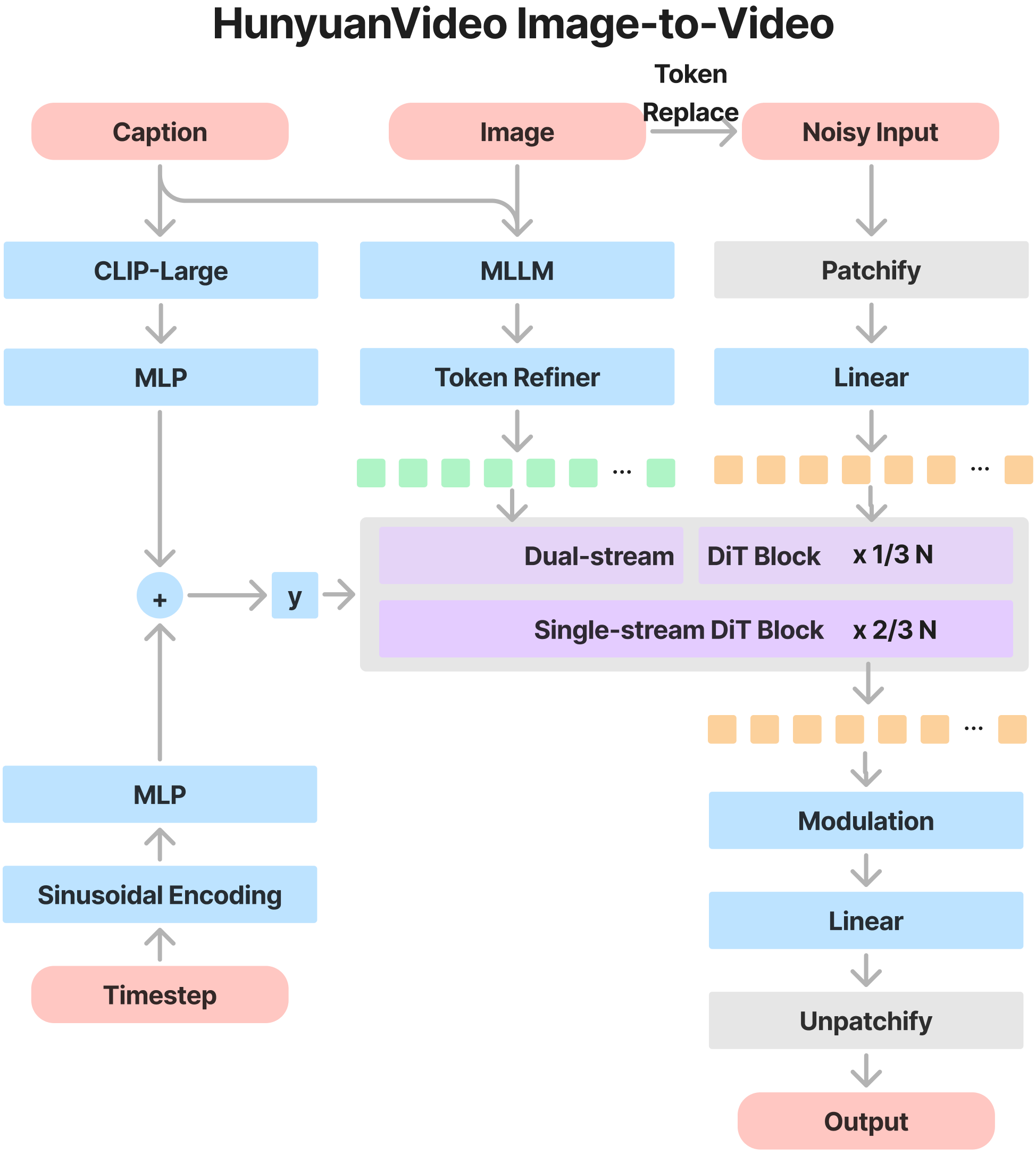}
    \fi
    \caption{HunyuanVideo-I2V Diffusion Backbone.}
    \label{fig:i2v_backbone}
\end{figure}
\begin{figure}[t]
    \centering
    \ifhq
    \includegraphics[width=\linewidth]{hqfigures/i2v_pretraining_result.png}
    \else
    \includegraphics[width=\linewidth]{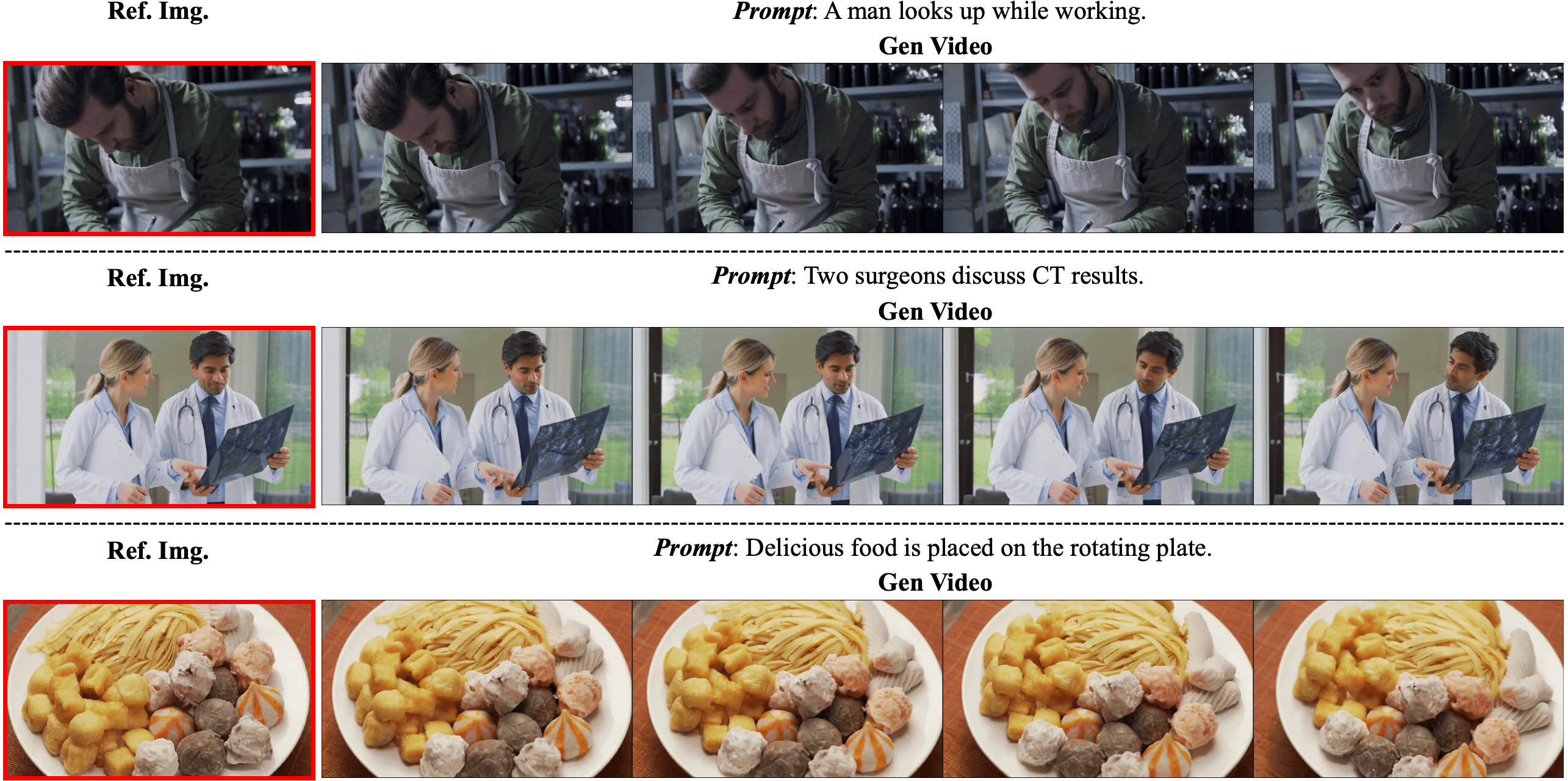}
    \fi
    \caption{Sample results of the I2V pre-training model.}
    \label{fig:i2v_result}
\end{figure}

Image-to-video (I2V) task is a common application in video generation tasks. It usually means that given an image and a caption, the model uses this image as the first frame to generate a video that matches the caption.
Although the na\"{\i}ve \nameofmethod{} is a text-to-video (T2V) model, it can be easily extended to an I2V model.
As shown in Fig. \ref{fig:i2v_backbone}, the I2V model employs a token replace technique to assist the model in more accurately reconstructing the original image information in its output.
The reference image latent is directly used as the first frame latent, and the corresponding timestep is set to 0. The processing of other frame latents is consistent with T2V training. 
To enhance the model's capability to comprehend the semantics of the input image and to more effectively integrate the information from both the image and the caption, the I2V model incorporates a semantic image injection module. It first inputs the image into the MLLM model to obtain the semantic image token and then concatenates these tokens into the video latent token for full-attention calculation.
We pre-train the I2V model on the same data as the T2V model, and the results are shown in Fig. \ref{fig:i2v_result}.

%% file: applications/human_i2v.tex
\subsubsection{Downstream Task Fine-tuning: Portrait Image-to-Video Generation}
We perform supervised finetuning of our I2V model on two million portrait videos to enhance human's motion and overall aesthetics. In addition to the standard data filtering pipeline described in section \ref{sec:data}, we also apply face and body detectors to filter out the training videos which have more than five persons. We also remove the videos in which the main subjects are small. Finally, the rest videos will be manually inspected to obtain the final high-quality portrait training dataset.

\begin{figure}[h]
    \centering
    \includegraphics[width=\linewidth]{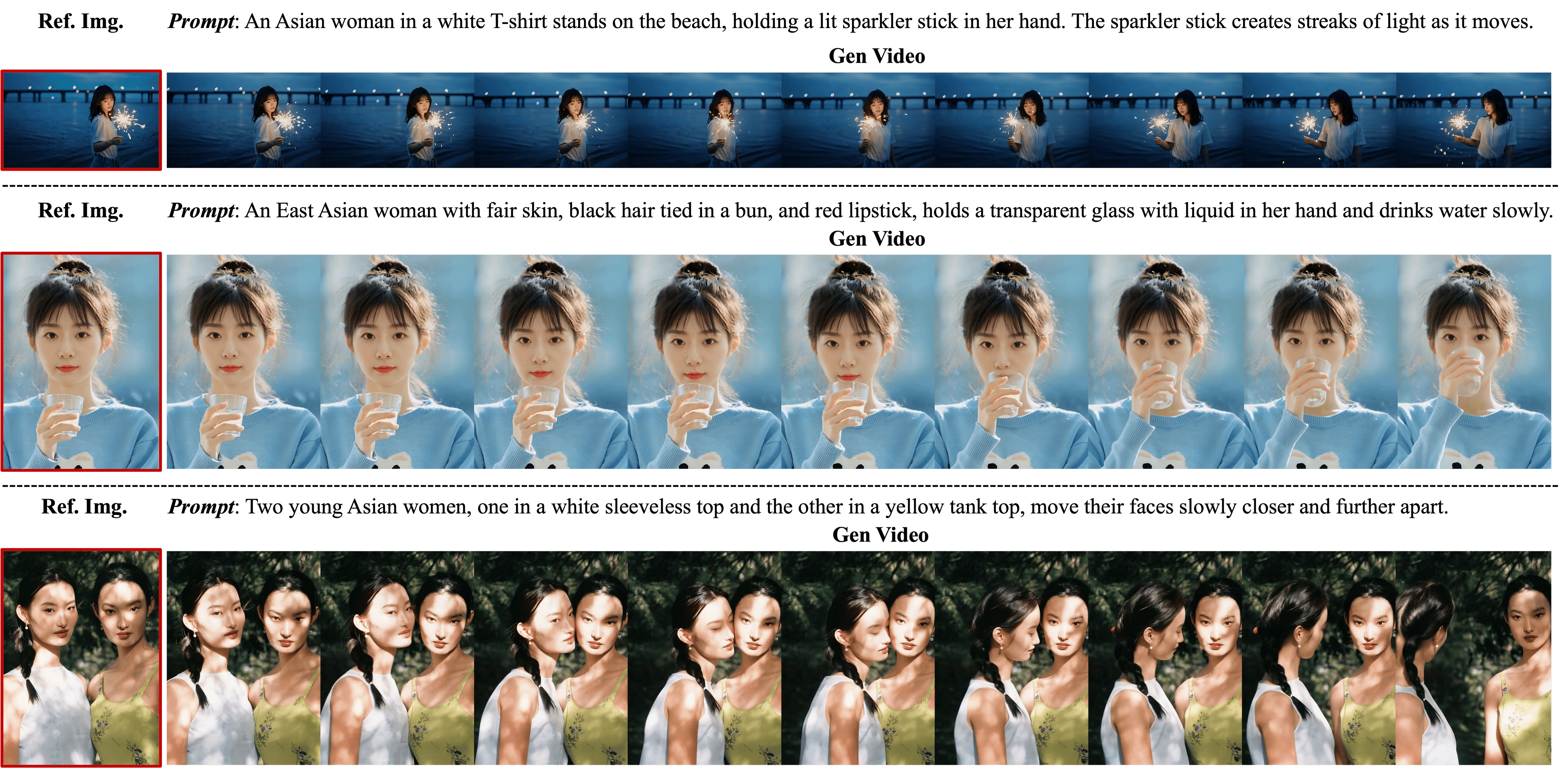}
    \caption{Sample results of our portrait I2V model.}
    \label{fig:portrait_i2v}
\end{figure}

Regarding training, we adopt a progressive fine-tuning strategy, gradually unfreezing the model parameters of the respective layers while keeping the rest frozen during finetuning. This approach allows the model to achieves high performance in the portrait domain without compromising much of its inherent generalization ability, guaranteeing commendable performance in natural landscapes, animals, and plants domains. Moreover, our model also supports video interpolation by using the first and last frames as conditions. We randomly drop the text conditions at certain probability during training to enhance the model's performance. Some demo results are shown in Fig. \ref{fig:portrait_i2v}.

%% file: applications/audio.tex
\subsection{Avatar animation}

\begin{figure}[h]
    \centering
    \includegraphics[width=\linewidth]{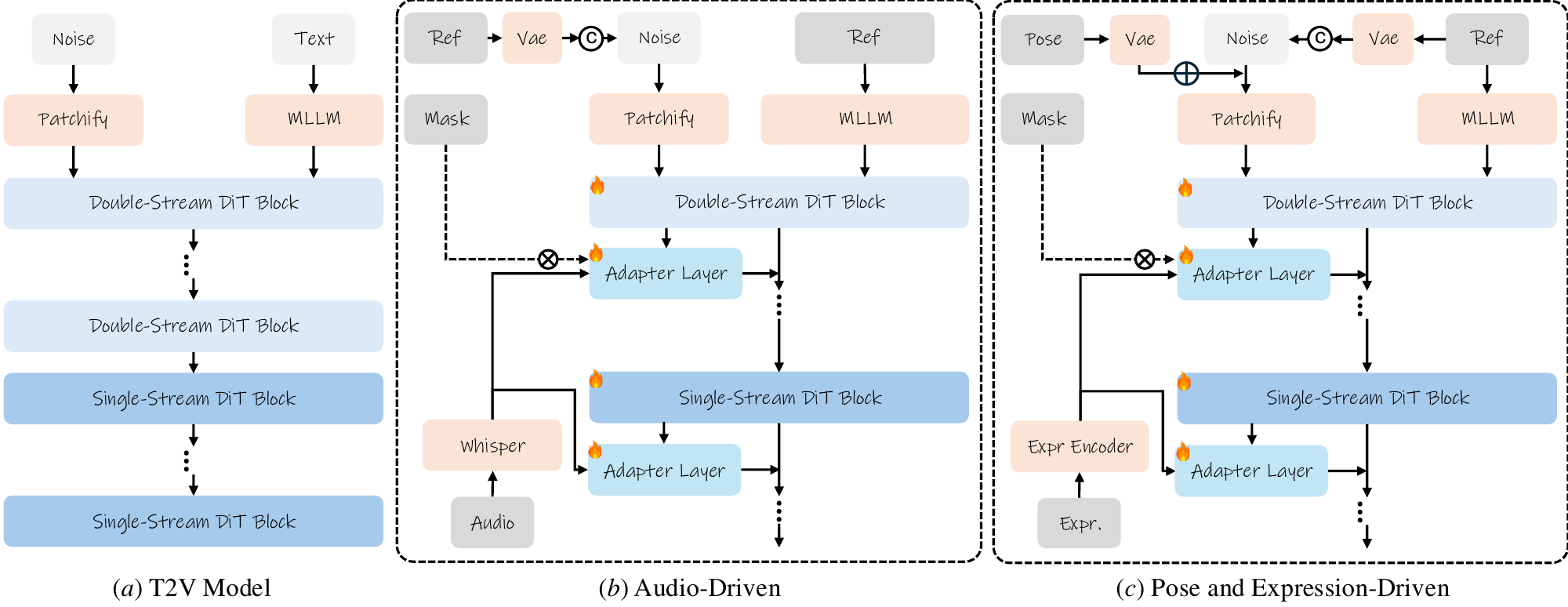}
    \caption{\textbf{Overview of Avatar Animation built on top of \nameofmethod}. We adopt 3D VAE to encode and inject reference and pose condition, and use additional cross-attention layers to inject audio and expression signals. Masks are employed to explicitly guide where they are affecting.}
    \label{fig:application-method}
\end{figure}

{\nameofmethod} empowers controllable avatar animation in various aspects. It enables animating characters using explicit driving signals(e.g., speech signals, expression templates, and pose templates). In addition, it also integrates the implicit driving paradigm using text prompts. Fig. \ref{fig:application-method} shows how we leverage the power of {\nameofmethod} to animate characters from multi-modal conditions. To maintain strict appearance consistency, we modify the {\nameofmethod} architecture by inserting latent of reference image as strong guidance. As shown in Fig. \ref{fig:application-method} (b, c), we encode reference image using 3DVAE obaining $z_{\rm ref} \in \mathbb{R}^{1 \times c \times h \times w}$, where $c = 16$. Then we repeat it $t$ times along temporal dimension and concatenate with $z_t$ in channel dimension to get the modified noise input $\hat{z}_t \in \mathbb{R}^{t \times 2c \times h \times w}$. To achieve controllable animation, various adapters are employed. We describe them in following.

\subsubsection{Upper-Body Talking Avatar Generation}

In recent years, audio-driven digital human algorithms have made significant progress, especially in the performance of the talking head. Early algorithms, such as loopy~\cite{ye2024mimic}, emo~\cite{tian2024emo}, and hallo~\cite{xu2024hallo}, mainly focused on the head area, driving the digital human's facial expressions and lip shapes by analyzing audio signals. Even earlier algorithms, like wav2lip~\cite{prajwal2020lip} and DINet~\cite{zhang2023dinet}, concentrated on modifying the mouth region in the input video to achieve lip shape consistency with the audio. However, these algorithms are usually limited to the head area, neglecting other parts of the body. To achieve a more natural and vivid digital human performance, we propose an audio-driven algorithm extended to the upper body. In this algorithm, the digital human not only synchronizes facial expressions and lip shapes with the audio while speaking but also moves the body rhythmically with the audio.

\paragraph{Audio-Driven}
Based on the input audio signal, our model can adaptively predict the digital human's facial expressions and posture action information . This allows the driven character to speak with emotion and expression, enhancing the digital human's expressiveness and realism. 
As shown in  Fig. \ref{fig:application-method} (b), for the single audio signal-driven part, the audio passes through the whisper feature extraction module to obtain audio features, which are then injected into the main network in a cross-attention manner. It should be noted that the injection process will be multiplied by the face-mask to control the audio's effect area. While enhancing the head and shoulder control ability, it will also greatly reduce the probability of body deformation. To obtain more lively head movements, head pose motion parameters and expression motion parameters are introduced and added to the time step in an embedding manner. During training, the head motion parameters are given by the variance of the nose tip keypoint sequence, and the expression parameters are given by the variance of the facial keypoints. 

\begin{figure}
    \centering
    \ifhq
    \includegraphics[width=\linewidth]{hqapplications/audio.png}
    \else
    \includegraphics[width=\linewidth]{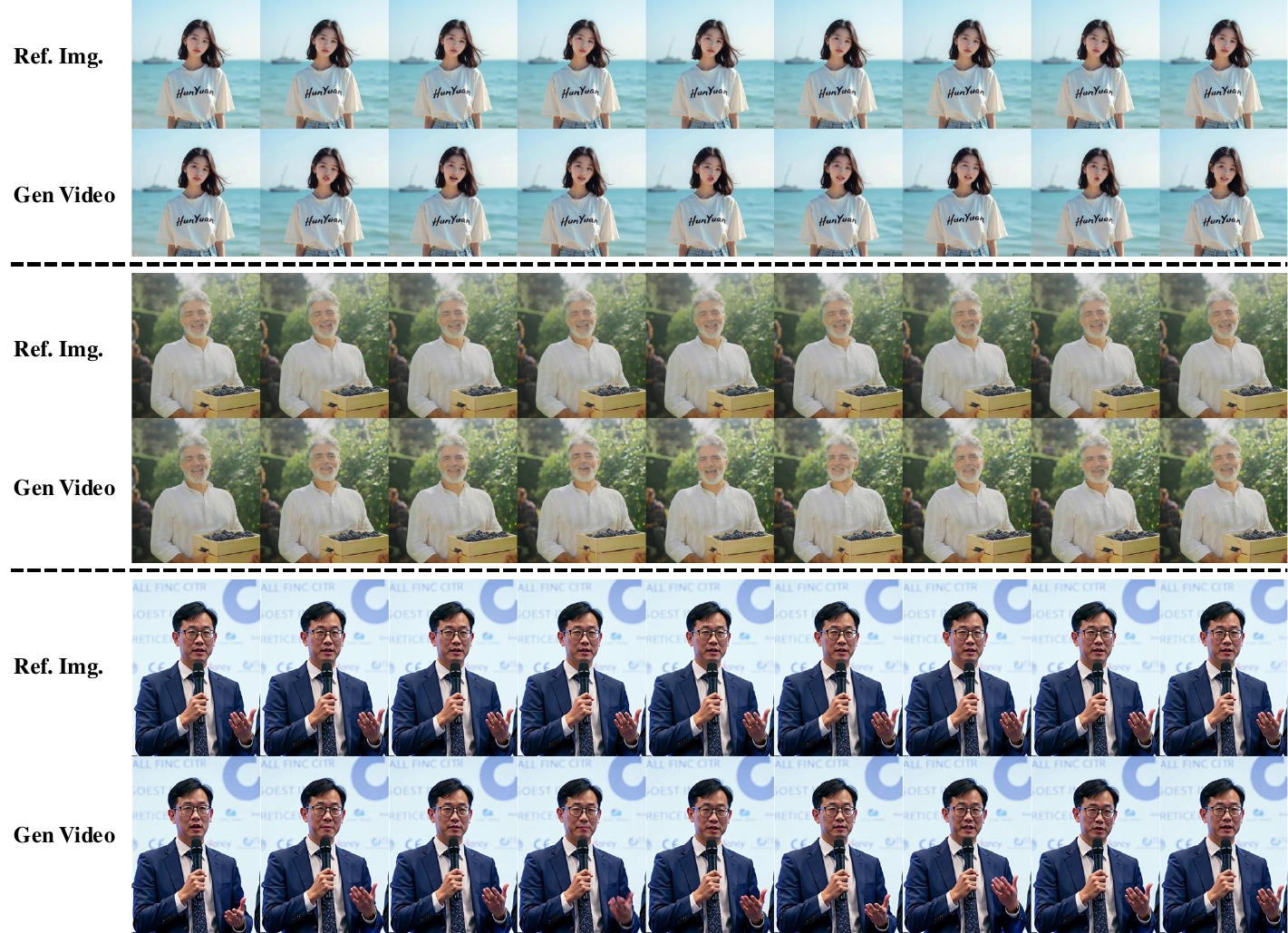}
    \fi
    \caption{\textbf{Audio-Driven}. {\nameofmethod} can generate vivid talking avatar videos.}
    \label{fig:application-audio}
\end{figure}

\begin{figure}
    \centering
    \ifhq
    \includegraphics[width=\linewidth]{hqapplications/pose.png}
    \else
    \includegraphics[width=\linewidth]{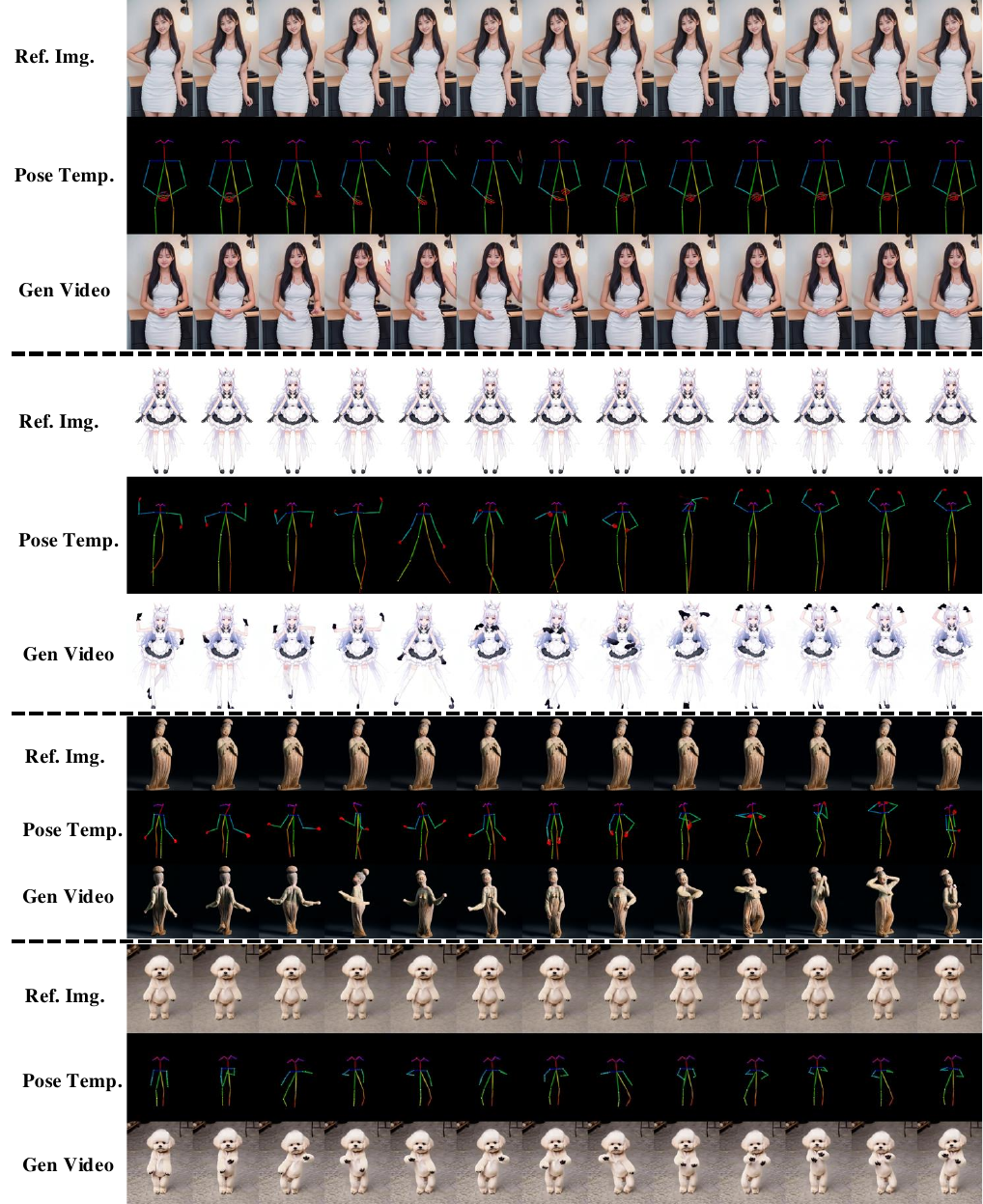}
    \fi
    \caption{\textbf{Pose-Driven}. {\nameofmethod} can animate wide variety of characters with high quality and appearance consistency under various poses.}
    \label{fig:application-pose}
\end{figure}

\begin{figure}
    \centering
    \ifhq
    \includegraphics[width=\linewidth]{hqapplications/expr.png}
    \else
    \includegraphics[width=\linewidth]{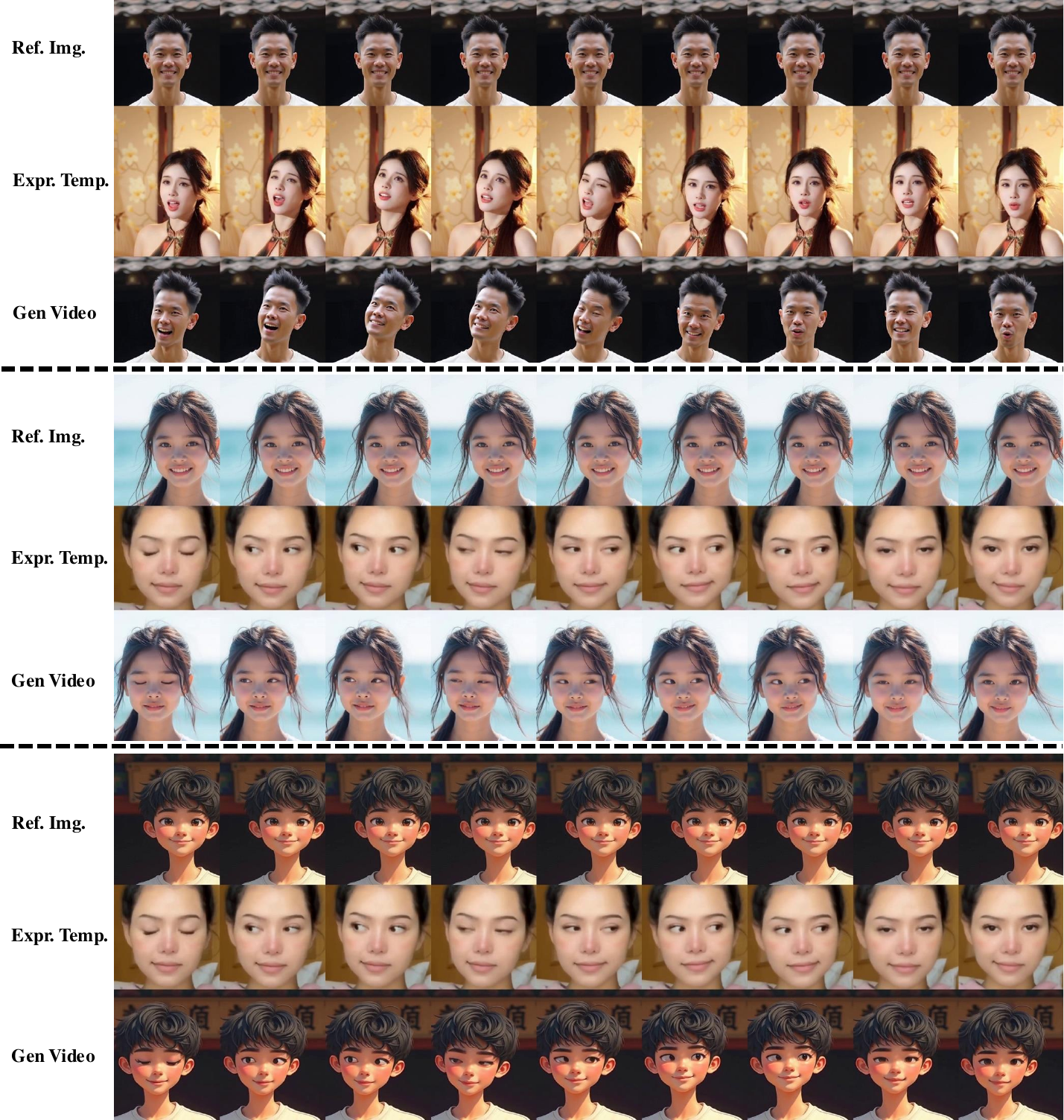}
    \fi
    \caption{\textbf{Expression-Driven}. {\nameofmethod} can accurately control facial movements of wide-variety of avatar styles.}
    \label{fig:application-expr}
\end{figure}

\begin{figure}[h]
    \centering
    \ifhq
    \includegraphics[width=\linewidth]{hqapplications/pose-expr.png}
    \else
    \includegraphics[width=\linewidth]{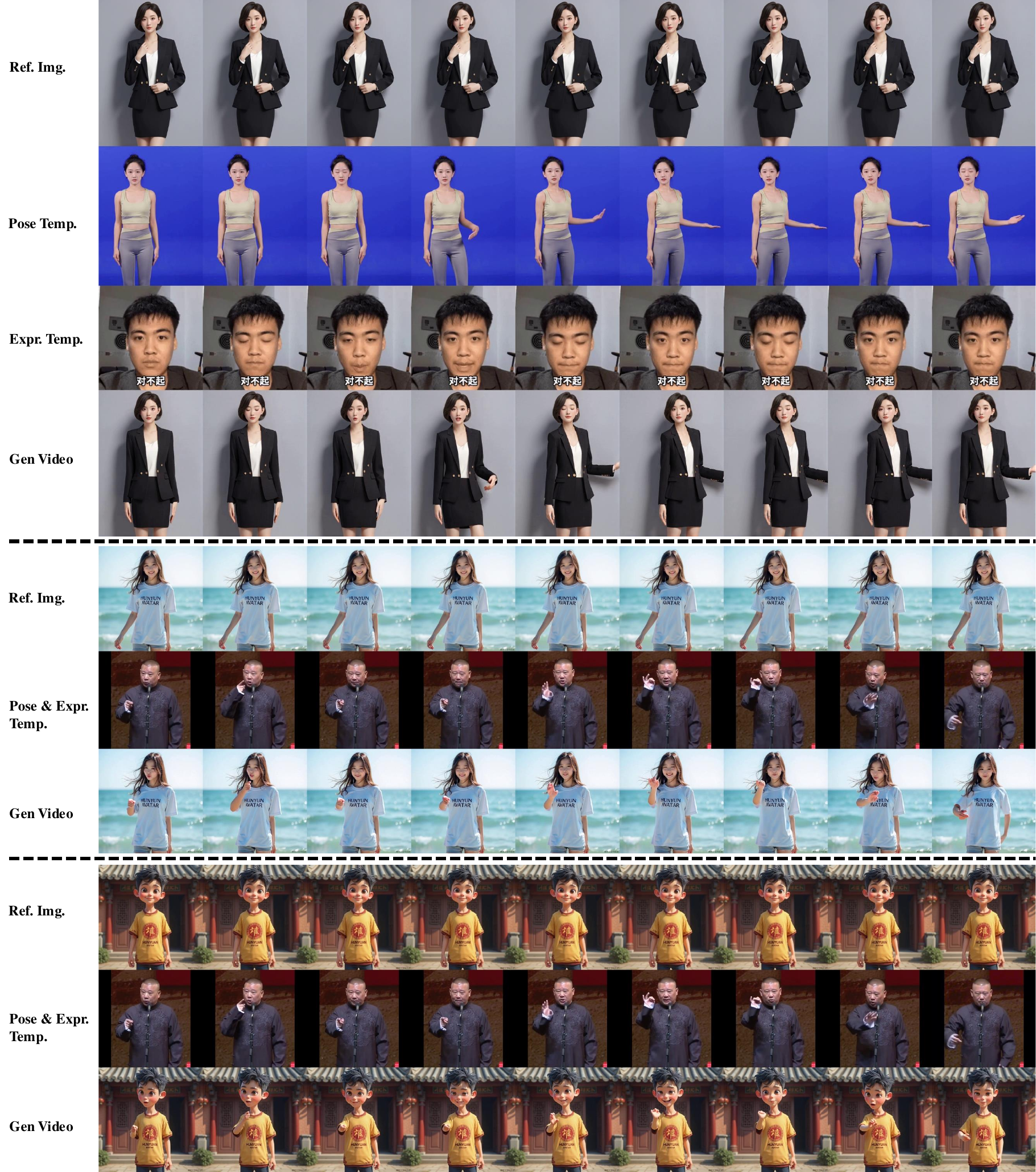}
    \fi
    \caption{\textbf{Hybrid Condition-Driven}. {\nameofmethod} supports full control with multiple driving sources across various avatar characters.}
    \label{fig:application-pose-expr}
\end{figure}

%% file: applications/expr_pose.tex
\subsubsection{Fully-Controlled Whole-Body Avatar Generation} Controlling digital character's motion and expression explicitly has been a long-standing problem in both academia and industry, and recent advancement of diffusion models paved the first step to realistic avatar animation. However, current avatar animation solutions suffer from partial controllability due to limited capability of foundation video generation model. We demonstrate that a stronger T2V model boosts the avatar video generation to fully-controllable stage. We show how {\nameofmethod} serves as strong foundation with limited modifications to extent general T2V model to fully-controllable avatar generation model in Fig. \ref{fig:application-method} (c).

\paragraph{Pose-Driven}
We can control the digital character's body movements explicitly using pose templates. We use Dwpose~\cite{yang2023effective} to detect skeletal video from any source video, and use 3DVAE to transform it to latent space as $z_{\rm pose}$. We argue that this eases the fine-tuning process because both input and driving videos are in image representation, and are encoded with shared VAE, resulting same latent space. We then inject the driving signals to the model by element-wise add as $\hat{z}_t + z_{\rm pose}$. Note that $\hat{z}_t$ contains the appearance information of reference image. We use full-parameters finetune with pretrained T2V weights as initialization. 

\paragraph{Expression-Driven}
We can also control the facial expressions of digital character using implicit expression representations. Although facial landmarks are widely adopted in this area~\cite{ma2024follow, chen2024echomimic}, we argue using landmarks brings ID leak due to cross-ID misalignment. Instead, we use implicit representations as driving signals for their ID and expression disentanglement capabilities. In this work, we use VASA~\cite{xu2024vasa} as expression extractor. As shown in Fig. \ref{fig:application-method} (c), we adopt a light-weight expression encoder to transform the expression representation to token sequence in latent space as $z_{\rm exp} \in \mathbb{R}^{t \times n \times c}$, where $n$ is the number of tokens per frame. Typically, we set $n = 16$. Unlike pose condition, we inject $z_{\rm exp}$ using cross-attention because $\hat{z}_t$ and $z_{\rm exp}$ are not naturally aligned in spatial aspect. We add cross-attention layer ${\rm Attn_{exp}}(q,k,v)$ every $K$ double and single-stream DiT layers to inject expression latent. Denote the hidden states after $i$-th DiT layer as $h_{i}$, the injection of expression $z_{\rm exp}$ to $h_{i}$ could be derived as: $h_{i} + {\rm Attn_{exp}} (h_i, z_{\rm exp}, z_{\rm exp}) \ast \mathcal{M}_{\rm face}$, where $\mathcal{M}_{\rm face}$ is the face region mask that guides where $z_{\rm exp}$ should be applied at, and $\ast$ stands for element-wise multiplication. Also, full-parameters tuning strategy is adopted.

\paragraph{Hybrid Condition Driven}
Combining both pose and expression driven strategies derives hybrid control approach. In this scenario, the body motion is controlled by explicit skeletal pose sequence, and the facial expression is determined by implicit expression representation. We jointly fine-tune T2V modules and added modules in an end-to-end fasion. During inference, the body motion and facial motion could be controlled by separate driving signals, empowering richer editability.

%% file: applications/demo.tex
\subsection{Application Demos} We present extensive results of avatar animations to show the superiority and potential of bringing avatar animation empowered by {\nameofmethod} to next generation.

\paragraph{Audio-Driven}Fig. \ref{fig:application-audio} shows that {\nameofmethod} serves as a strong foundation model for audio-driven avatar animation, which can synthesize vivid and high-fidelity videos. We summarize the superiority of our method in three folds:

\begin{itemize}
    \item \textbf{Upper-body Animation.} Our method can drive not only portrait characters, but also upper-body avatar images, enlarging its range of application scenarios. 
    \item \textbf{Dynamic Scene Modelling.} Our method can generate videos with vivid and realistic background motion, such as the wave undulation, crowd movement, and breeze stirring leaves. 
    \item \textbf{Vivid Avatar Movements.} Our method is able to animate the character talking while gesturing vividly with audio solely.
\end{itemize}

\paragraph{Pose-Driven}We also show that {\nameofmethod} boosts the performance of pose-driven animation largely in many aspects in Fig. \ref{fig:application-pose}:

\begin{itemize}
    \item \textbf{High ID-Consistency.} Our method maintains the ID-consistency well over the frames even with large poses, making it face-swapping free, thereby, could be used as real end-to-end animation solution.
    \item \textbf{Following Complex Poses Accurately.} Our method is able to handle very complex poses such as turning around and hands crossed.
    \item \textbf{High Motion Quality.} Our method has remarkable capability in dynamic modelling. For instance, the results show promising performance in terms of garment dynamics and texture consistency.
    \item \textbf{Generalizability.} Our method presents surprisingly high generalizability. It can animate wide variety of avatar images, such as real human, anime, pottery figurine, and even animals. 
\end{itemize}

\paragraph{Expression-Driven}Fig. \ref{fig:application-expr} presents how {\nameofmethod} enhances the portrait expression animating in three folds:

\begin{itemize}
    \item \textbf{Exaggerated Expression.} Our method is able to animate given portrait to mimic any facial movements even with large poses and exaggerated expressions.
    \item \textbf{Mimicing Eye Gaze Accurately.} We can control the portraits' eye movements acurately given any expression template, even with extreme and large eye balls movements.
    \item \textbf{Generalizability.} Our method has high generalizability. It can animate not only real human portraits, but also anime or CGI characters.
\end{itemize}

\paragraph{Hybrid-Driven} Lastly, we show that hybrid condition control reveals the potential of fully controllable and editable avatars in Fig. \ref{fig:application-pose-expr}. We highlight the superiority as follow:

\begin{itemize}
    \item \textbf{Hybrid Condition Control.} For the first time, our method is able to conduct full control over body and facial motions with siloed or multiple signals, paving the route from demo to applications for avatar animation.
    \item \textbf{Half-body Animation.} Our method supports upper-body full control, enabling rich editability while maintaining high quality and fidelity.
    \item \textbf{Generalizability.} Our method generalize to both real human images and CGI characters. 
\end{itemize}



%% file: tax/contributors.tex
\newpage
\section*{Project Contributors}

\begin{itemize}

  \item \textbf{Project Sponsors:} Jie Jiang, Yuhong Liu, Di Wang, Yong Yang 
    \item \textbf{Project Leaders:} Caesar Zhong, Hongfa Wang, Dax Zhou, Songtao Liu, Qinglin Lu, Yangyu Tao
    \item \textbf{Core Contributors:}
    \begin{itemize}
        \item \textbf{Infrastructure:} Rox Min, Jinbao Xue, Yuanbo Peng, Fang Yang, Shuai Li, Weiyan Wang, Kai Wang
        \item \textbf{Data \& Recaptioning:} Zuozhuo Dai, Xin Li, Jin Zhou, Junkun Yuan, Hao Tan, Xinchi Deng, Zhiyu He, Duojun Huang, Andong Wang, Mengyang Liu, Pengyu Li
        \item \textbf{VAE \& Model Distillation:} Bo Wu, Rox Min, Changlin Li, Jiawang Bai, Yang Li, Jianbing Wu 
        \item \textbf{Algorithm \& Model Architecture \& Pre-training:} Weijie Kong, Qi Tian, Jianwei Zhang, Zijian Zhang, Kathrina Wu, Jiangfeng Xiong, Yanxin Long
        \item \textbf{Downstream Tasks:} Jacob Song, Jin Zhou, Yutao Cui, Aladdin Wang, Wenqing Yu, Zhiyong Xu, Zixiang Zhou, Zhentao Yu, Yi Chen, Hongmei Wang, Zunnan Xu, Joey Wang, Qin Lin
    \end{itemize}
    \item \textbf{Contributors:} Jihong Zhang, Meng Chen, Jianchen Zhu, Winston Hu, Yongming Rao, Kai Liu, Lifei Xu, Sihuan Lin, Yifu Sun, Shirui Huang, Lin Niu, Shisheng Huang, Yongjun Deng, Kaibo Cao, Xuan Yang, Hao Zhang, Jiaxin Lin, Chao Zhang,  Fei You, Yuanbin Chen, Yuhui Hu, Liang Dong Zheng Fang, Dian Jiao, Zhijiang Xu, Xuhua Ren, Bing Ma, Jiaxiang Cheng, Wenyue Li, Kai Yu, Tianxiang Zheng
\end{itemize}